\pdfoutput=1

\documentclass[11pt]{article}

\usepackage{longtable} %
\usepackage{array} %
\usepackage{caption} %

\usepackage[dvipsnames]{xcolor}
\usepackage{authblk}
\usepackage{graphicx}
\usepackage{xurl}
\usepackage{enumitem}
\usepackage{xspace}
\usepackage{xparse}
\usepackage{float}

\DeclareDocumentCommand\newstep{o}{%
\item\IfNoValueTF{#1}{}{#1 \textendash\xspace}}
\newlist{steps}{enumerate}{1}
\setlist[steps]{label=\textit{RQ \arabic*:},leftmargin=*}

\usepackage[final]{acl}

\usepackage{times}
\usepackage{latexsym}
\usepackage{colortbl}
\usepackage{amssymb}
\usepackage{tablefootnote}
\usepackage{amsmath}
\usepackage{hyperref}
\usepackage{cleveref}
\crefrangelabelformat{section}{#3#1#4--#5\crefstripprefix{#1}{#2}#6}
\usepackage{subcaption}
\usepackage{dblfloatfix}

\usepackage{ragged2e}
\usepackage{tcolorbox}

\usepackage{listings}
\definecolor{codegreen}{rgb}{0,0.6,0}
\definecolor{codegray}{rgb}{0.5,0.5,0.5}
\definecolor{codepurple}{rgb}{0.58,0,0.82}
\definecolor{backcolour}{rgb}{0.95,0.95,0.92}

\lstdefinestyle{mystyle}{
    backgroundcolor=\color{backcolour},   
    commentstyle=\color{codegreen},
    keywordstyle=\color{magenta},
    numberstyle=\tiny\color{codegray},
    stringstyle=\color{codepurple},
    basicstyle=\ttfamily\footnotesize,
    breakatwhitespace=false,         
    breaklines=true,                 
    captionpos=b,                    
    keepspaces=true,                 
    numbers=left,                    
    numbersep=5pt,                  
    showspaces=false,                
    showstringspaces=false,
    showtabs=false,                  
    tabsize=2
}

\usepackage[T1]{fontenc}

\usepackage[utf8]{inputenc}

\usepackage{booktabs}
\usepackage{multirow}
\usepackage{makecell} 
\usepackage{arydshln}

\usepackage{tikz}
\newcommand{\overcurvedleftarrow}[1]{%
  \begin{tikzpicture}[baseline=(X.base)]
    \node (X) {$#1$};
    \draw[->, thick, bend left=-20] 
      ([yshift=1ex, xshift=-0.5ex]X.east) to ([yshift=1ex, xshift=0.5ex]X.west);
  \end{tikzpicture}%
}

\newcommand{\undercurvedrightarrow}[1]{%
  \begin{tikzpicture}[baseline=(X.base)]
    \node (X) {$#1$};
    \draw[->, thick, bend right=20] 
      ([yshift=-1ex, xshift=0.5ex]X.west) to ([yshift=-1ex, xshift=-0.5ex]X.east);
  \end{tikzpicture}%
}

\usepackage{microtype}

\usepackage{cuted}

\usepackage{inconsolata}

\newcommand{\theoremaug}{\textsc{ReasonAug}\xspace}
\newcommand{\longembed}{\textsc{LongEmbed}\xspace}
\newcommand{\modernbert}{ModernBERT\xspace}
\newcommand{\bright}{\textsc{Bright}\xspace}
\newcommand{\followir}{\textsc{FollowIR}\xspace}
\newcommand{\msmarco}{MS MARCO\xspace}
\newcommand{\dream}{\textsc{Dream}\xspace}
\newcommand{\ours}{\textsc{DiffEmbed}\xspace}

\newcommand{\eg}{\hbox{\emph{e.g.,}}\xspace}
\newcommand{\ie}{\hbox{\emph{i.e.,}}\xspace}

\newcolumntype{L}[1]{>{\raggedright\let\newline\\\arraybackslash\hspace{0pt}}m{#1}}
\newcolumntype{C}[1]{>{\centering\let\newline\\\arraybackslash\hspace{0pt}}m{#1}}
\newcolumntype{R}[1]{>{\raggedleft\let\newline\\\arraybackslash\hspace{0pt}}m{#1}}

\usepackage{tabularx}

\definecolor{lightgreen}{RGB}{230,255,230}
\definecolor{lightblue}{RGB}{230,230,255}
\definecolor{lightred}{RGB}{255,230,230}
\definecolor{lightgrey}{RGB}{245,245,245}

\definecolor{YaleYellow}{RGB}{179, 176, 4} %
\definecolor{NYUPurple}{RGB}{134, 1, 175}  %
\definecolor{NTUBlue}{RGB}{2,2,200} 
\definecolor{Alibaba}{RGB}{255, 106, 0}%
\definecolor{Center}{RGB}{0, 128, 0}%

\title{Diffusion vs. Autoregressive Language Models:\\A Text Embedding Perspective}

\author{
\bf{Siyue Zhang$^{\hspace{.1em}{\textcolor{NTUBlue}{\boldsymbol{N}}}}$$^{\hspace{.1em}{\textcolor{Alibaba}{\boldsymbol{A}}}}$ \quad
Yilun Zhao$^{\hspace{.1em}\textcolor{YaleYellow}{\boldsymbol{Y}}}$ \quad
Liyuan Geng$^{\hspace{.1em}\textcolor{NYUPurple}{\boldsymbol{S}}}$
\quad
Arman Cohan$^{\hspace{.1em}\textcolor{YaleYellow} {\boldsymbol{Y}}}$} 
\vspace{-9pt}\\
\bf{
Anh Tuan Luu$^{\hspace{.1em}\textcolor{NTUBlue}{\boldsymbol{N}}}$ \quad
Chen Zhao$^{\hspace{.1em}\textcolor{NYUPurple}{\boldsymbol{S}}}$
$^{\hspace{.1em}\textcolor{Center}{\boldsymbol{C}}}$
}
\vspace{9pt}\\
$^{\textcolor{NTUBlue}{\boldsymbol{N}}}$Nanyang Technological University \quad
$^{\textcolor{YaleYellow}{\boldsymbol{Y}}}$Yale University \quad
$^{\textcolor{NYUPurple}{\boldsymbol{S}}}$NYU Shanghai \quad
\\
$^{\textcolor{Alibaba}{\boldsymbol{A}}}$Alibaba-NTU Singapore Joint Research Institute 
\\
$^{\textcolor{Center}{\boldsymbol{C}}}$Center for Data Science, New York University}

\begin{document}

\maketitle

\begin{abstract}

Large language model (LLM)-based embedding models, benefiting from large scale pre-training and post-training, have begun to surpass BERT and T5-based models on general-purpose text embedding tasks such as document retrieval. However, a fundamental limitation of LLM embeddings lies in the unidirectional attention used during autoregressive pre-training, which misaligns with the bidirectional nature of text embedding tasks. To this end, We propose  adopting diffusion language models for text embeddings, motivated by their inherent bidirectional architecture and recent success in matching or surpassing LLMs especially on reasoning tasks. We present the first systematic study of the diffusion language embedding model, which outperforms the LLM-based embedding model by 20\% on long-document retrieval, 8\% on reasoning-intensive retrieval, 2\% on instruction-following retrieval, and achieve competitive performance on traditional text embedding benchmarks. Our analysis verifies that bidirectional attention is crucial for encoding global context in long and complex text.\footnote{Our code and data are available at \url{https://github/anonymous}.}

\end{abstract}

\begin{figure}[t!]
    \centering

    \begin{subfigure}{0.38\textwidth}
        \includegraphics[width=\linewidth]{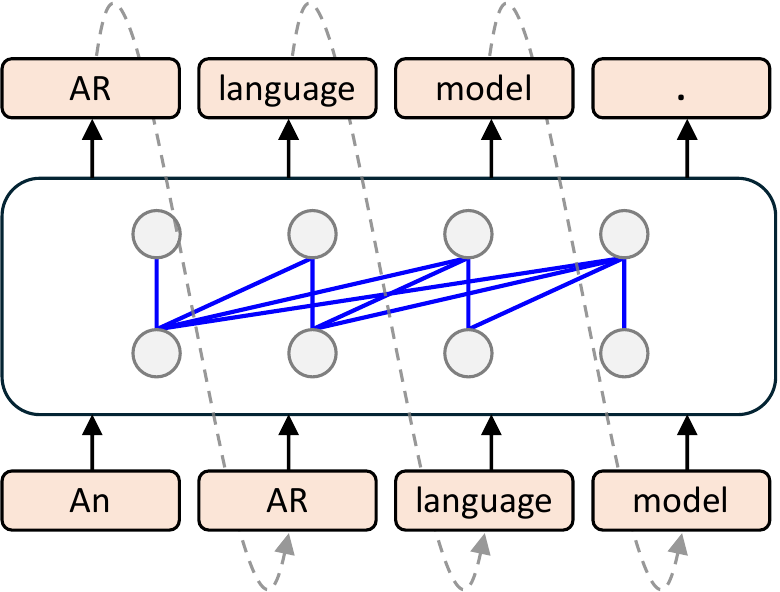}
        \caption{Autoregressive Modelling}
        \label{llm_arc}
    \end{subfigure}
    
    \vspace{0.5em} %

    \begin{subfigure}{0.38\textwidth}
        \includegraphics[width=\linewidth]{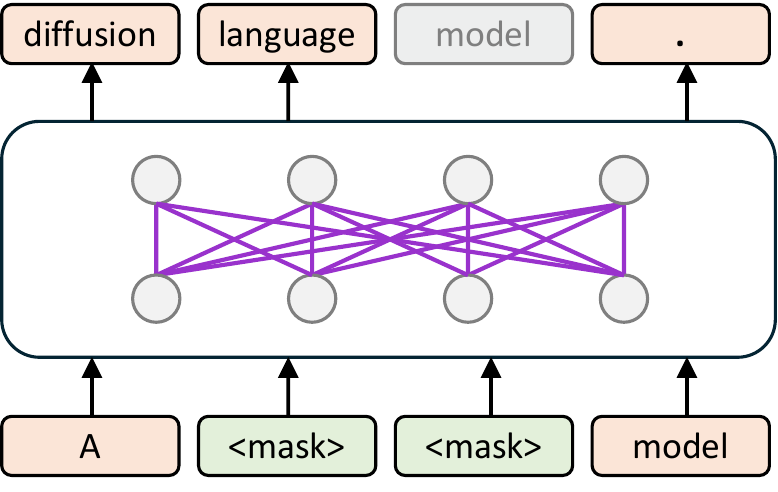}
        \caption{Diffusion Modelling in \dream}
        \label{dream_arc}
    \end{subfigure}
    
    \vspace{0.5em} %

    \begin{subfigure}{0.48\textwidth}
        \includegraphics[width=\linewidth]{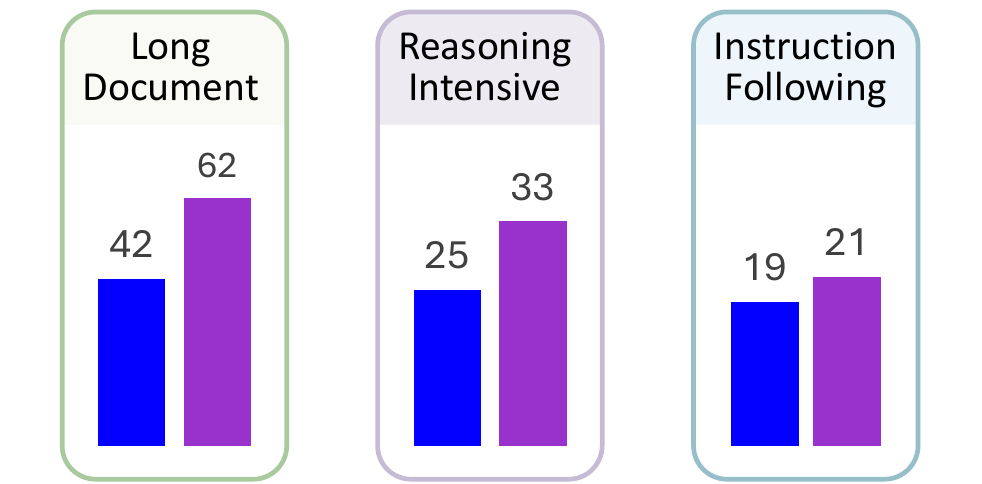}
        \caption{Llama3 + LLM2Vec (blue) vs. \dream (purple)}
    \end{subfigure}

    \caption{\textbf{(a)} Unidirectional attention in Autoregressive LM. \textbf{(b)} Bidirectional attention in Diffusion LM, \ie \dream \citep{dream}. \textbf{(c)} Retrieval performance comparison between the diffusion embedding model and the LLM embedding model enhanced with LLM2Vec adaptation \citep{llm2vec}.
    }
    \label{fig:three_rows}
\end{figure}

\section{Introduction}

Learning text embeddings is a fundamental NLP problem that supports a wide range of downstream applications such as retrieval-augmented generation (RAG). Traditionally,  text embedding models \cite{dpr,contriever,SBERT} have been  trained using contrastive learning on top of pre-trained bidirectional language models such as BERT \citep{bert} and T5 \citep{t5}. 
More recently, several studies \citep{repllama,e5,bge_gemma,nv_embed,gritlm} have adapted decoder-only large language models (LLMs), achieving notable improvements on embedding benchmarks such as MTEB \citep{mteb}.

Despite their strong empirical performance, LLM-based embedding models struggle to capture global context due to the causal attention structure used during large-scale training \citep{echo}. As illustrated in \Cref{llm_arc}, contextualized token embeddings—the final-layer hidden states at each position—are computed without access to future tokens. This leads to a mismatch between the unidirectional nature of LLM training paradigm and the bidirectional context understanding required for text embedding tasks. Echo Embeddings \citep{echo} attempts to address this issue by duplicating the input and extracting embeddings from the second copy, though this adds inference overhead, particularly for long texts. LLM2Vec \citep{llm2vec} adapts LLMs with bidirectional attention through continue training, but is limited with scale. %

We present \ours, a novel approach that leverages diffusion language models (LMs) for text embedding. Recent diffusion LMs design discrete diffusion processes using forward masking and reverse unmasking within a bidirectional attention architecture.  Trained at scale similarly to autoregressive LLMs, these models have demonstrated competitive performance across a variety of tasks \citep{simple,llada,dream}.
\textit{Our hypothesis is that diffusion embeddings, when trained at scale with a bidirectional attention architecture, are better suited to capturing global context and thus achieve superior performance on long and complex documents.}

To test this hypothesis, we  evaluate \ours and LLM embedding models\footnote{For brevity, we refer to the autoregressive and diffusion LM-based embedding models as the LLM and diffusion embedding models, respectively.} over a wide range of tasks, including long-document retrieval, reasoning-intensive (logical) retrieval, instruction-following retrieval, and general text embedding tasks. As no existing training dataset is effective for the reasoning-intensive task, we develop a new training set \theoremaug using LLMs, which contains 10,896 pairs of logically related positives and unrelated hard negatives. These documents cover a range of domains, from mathematics and physics theorems  to code.

Our experimental results show that \ours, based on the state-of-the-art diffusion LM \dream-7B \citep{dream}, outperforms the Llama3-8B based model by 20\% on the long-document retrieval benchmark \longembed~\citep{longembed}, 8\% on the reasoning-intensive retrieval benchmark \bright~\citep{bright}, and 2\% on instruction-following retrieval benchmark \followir~\citep{followir}. Notably, with \theoremaug, \ours surpasses the state-of-the-art performance on TheoremQA tasks in \bright by 16.4\%. Our ablation study shows that attention in both directions (\overcurvedleftarrow{\text{causal}} and \undercurvedrightarrow{\text{reverse}}) is crucial for encoding long and complex documents. The \undercurvedrightarrow{\text{reverse}} attention has a greater impact in \ours than LLM-based models.

To summarize, our contributions include:

\begin{itemize}[leftmargin=*]
\item We propose to leverage the diffusion LMs for text embedding, which are trained at scale with a bidirectional architecture. This approach is intuitively motivated and avoids adaptations typically required for LLM embeddings \citep{llm2vec}.

\item To the best of our knowledge, we are the first to systematically evaluate diffusion embeddings, which demonstrate superior long-document encoding and logical reasoning capabilities, as well as competitive performance on general text embedding tasks.

\item We present \theoremaug, a new dataset which is constructed solely using LLMs and significantly improves the retrieval performance for logical documents, which requires intensive reasoning.
\end{itemize}

\section{Background}
\label{sec: beckground}

\paragraph{Text Embedding Tasks.} The goal of text embedding models is to map a sequence of tokens $x = x_1, ..., x_n$ to a vector $\phi(x)\in\mathbb{R}^d$ that preserves semantic similarity within a low-dimensional space. Embeddings are used in a wide range of downstream applications such as document retrieval, clustering, classification, among others \citep{mteb}. Specifically, in document retrieval, for each query embedding, document embeddings are ranked by semantic similarity; in clustering, embeddings form semantically coherent groups; and in classification, embeddings are directly mapped to class labels.

\paragraph{Text Embedding Models.} Text embedding models are typically adapted from pre-trained language model to leverage their contextual understanding of text. Embeddings are obtained from language model's final layer, where each input token $x_j$ at position $j$ is associated with a contextualized representation $\phi_j(x)$. To obtain a fixed-size sequence embedding, token representations are aggregated---commonly via mean pooling or by selecting the final token’s representation.

These embedding models are further fine-tuned through contrastive learning to produce more effective representations \citep{simcse, e5_weakly, bge_m3,jina}. Given a query $q$, a positive text $p^{+}$, and a set of negative text $p_{1}^{-}, \dots, p_{m}^{-}$ , the model is trained to increase the similarity between the query and the positive text while decreasing the similarity with negative text. This is achieved by optimizing the following objective \citep{dpr}:
\begin{equation}
\resizebox{0.95\hsize}{!}{$
\mathcal{L}(q, p^{+}, p_{1}^{-}, \dots, p_{m}^{-}) = \frac{e^{f(q, p^{+})}}{e^{f(q, p^{+})} + \sum_{j=1}^{m} e^{f(q, p^{-}_{j})}},
$}
\label{eqn:loss}
\end{equation}
where $f(q, p)$ denotes a similarity function, e.g., dot product.

For years, bidirectional LMs such as BERT \citep{bert} and T5 \citep{t5} have dominated as backbone for text embedding models. \modernbert \citep{modernbert} scales up BERT training with a longer sequence length. Recent works have made plentiful advances to adapt decoder-only LLMs for text embeddings such as Repllama \citep{repllama}, E5-Mistral \citep{e5}, and NV-Embed \citep{nv_embed}. To overcome the limitation of unidirectional attention in LLMs, LLM2Vec \citep{llm2vec} introduce the self-supervised training to adapt LLMs to use bidirectional attention for embedding tasks.

\paragraph{Diffusion Language Models.} 

Diffusion models \citep{sohl, scorebased} excel in generative tasks, especially in image and video generation. Extending these models to the discrete domain of natural language offers a promising direction for addressing key limitations of autoregressive LMs, including incoherent output \citep{coherence}, limited controllability \citep{control_issue}, and slow inference speed \citep{speculative}.

To apply diffusion models to text, one line of approaches transforms discrete text into a continuous latent space, applies a diffusion process and then decodes the output back into discrete text \citep{wu2023ardiffusion,tess}, while another line of approaches designs discrete diffusion processes with new forward and reverse dynamics tailored to discrete tokens \citep{austin,simple,sedd,llada}.
Specifically, for a model distribution \( p_\theta(x_0) \), the forward process gradually masks tokens in \( x_0 \) independently until yielding a fully masked sequence at \( t = 1 \). During the masking process \( t \in (0, 1) \), each token is masked with probability \( t \).
The reverse process recovers the data distribution by iteratively predicting masked tokens as \( t \) decreases from 1 to 0 \citep{llada}. A parametric mask predictor \( p_\theta(\cdot \mid x_t) \) takes a partially masked \( x_t \) as input and simultaneously predicts all masked tokens, denoted \( \mathbf{M} \). The model is trained using cross-entropy loss computed only over the masked tokens:

\begin{equation}
\resizebox{0.95\hsize}{!}{$ \displaystyle
\mathcal{L}(\theta) \triangleq -\mathbb{E}_{t, x_0, x_t} \left[ \frac{1}{t} \sum_{i=1}^{L} \mathbf{1}[x_t^i = \mathbf{M}] \log p_\theta(x_0^i \mid x_t) \right],
$}
\end{equation}

\noindent where $x_0$ is sampled from the training data, $t$ is sampled
uniformly from $[0, 1]$, and $x_t$ is sampled from the forward
process. The indicator function $\mathbf{1}[\cdot]$ ensures that the loss is computed only for masked tokens. Recent masked diffusion models such as \dream \citep{dream} scale to over one billion parameters and has shown strong performance on multiple math and code reasoning tasks.

\begin{figure}[t!]
    \centering
    \includegraphics[width=0.3\textwidth]{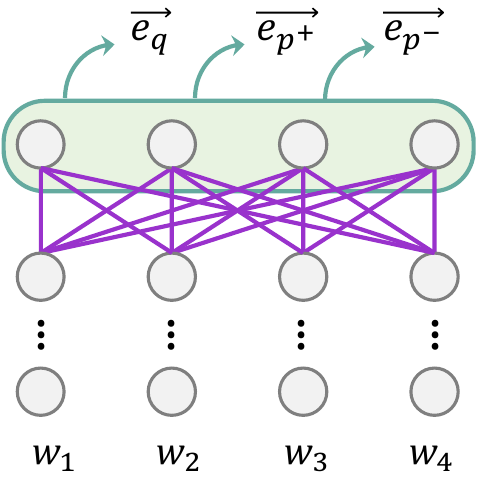}
    \captionsetup{justification=justified, singlelinecheck=false}  %
    \caption{Overview of \ours. Final-layer token representations from the backbone diffusion LM are mean-pooled to obtain text embeddings.}
    \label{fig:arc}
\end{figure}

\section{Diffusion Embedding Model}
\label{sec: method}

\begin{table*}[t!]
\centering
\renewcommand{\arraystretch}{1.1}
\captionsetup{justification=justified,singlelinecheck=false}
\begin{tabularx}{\textwidth}{
    >{\raggedright\arraybackslash}l 
    >{\centering\arraybackslash}X 
    >{\centering\arraybackslash}c
    >{\centering\arraybackslash}c
    >{\centering\arraybackslash}c
    >{\centering\arraybackslash}c
    >{\centering\arraybackslash}c
    >{\centering\arraybackslash}c
} 
\toprule
\multirow{2}{*}{\textbf{Training Task}} & \multirow{2}{*}{\textbf{Dataset}} & \multicolumn{2}{c}{\textbf{Tot. Number}} & \multicolumn{2}{c}{\textbf{Query Length}} & \multicolumn{2}{c}{\textbf{Doc. Length}}  \\
\cmidrule(lr){3-4} \cmidrule(lr){5-6} \cmidrule(lr){7-8}
& & \textbf{$\mathbb{N}$} & \textbf{N} & Avg. &  Std. & Avg. & Std. \\
\hline
Generalist embedding & Public E5   & 2M & 16k & 39 & 19 & 366 & 1062   \\
Theorem reasoning & \theoremaug &   10k & 10k & 230 & 244 & 508 & 427 
 \\
Instruction following & MS MARCO$^\dagger$ & 1M & 16k & 83 & 76 & 92 & 41   
  \\
\bottomrule
\end{tabularx}
\caption{Statistics of datasets used for training embedding models. For each dataset, we show the original number of samples ($\mathbb{N}$), the amount for training subsets (N), rounded average length of queries and documents (measured by GPT-2 tokenizer \citep{gpt2}), and standard deviation. $^\dagger$\msmarco with Instructions.}
\label{training_sets}
\end{table*}

In this section, we propose a new type of text embedding model, the Diffusion Embedding Model (\ours). In line with LLM-based text embedding models introduced in \Cref{sec: beckground}, \ours first extracts contextualized token representations from the backbone diffusion LM and aggregate them using mean pooling\footnote{As recommended by \citet{llm2vec, echo}, mean pooling offers better performance and robustness than alternatives like last-token pooling.} (see \Cref{fig:arc}). Then \ours learns effective text representations through contrastive learning on embedding tasks as in \Cref{eqn:loss}. Unlike LLMs, diffusion LMs are inherently bidirectional, removing the need for intermediate steps such as enabling bidirectional attention and continue pre-training \citep{nv_embed,llm2vec}.

The primary distinction between \ours and LLM-based embedding models lies in the mechanism by which embeddings are generated within the language models. LLM-based embeddings are obtained through large-scale pretraining using causal attention masking and a next-token prediction objective, which encourages the model to understand past context to predict future tokens \citep{gpt,llama3}. In contrast, diffusion LM embeddings are learned through a denoising objective, where the model is trained to recover noisy or corrupted inputs (\eg masked tokens) simultaneously \citep{llada,dream}. This process enables the embeddings to capture bidirectional semantics and potentially more robust representations of input text.

\section{Evaluating Text Embedding Models}
\label{setups}

To systematically evaluate \ours, we compare it with LLM embedding models on diverse set of tasks including long-document retrieval (\S\ref{sec:long}), reasoning-intensive retrieval (\S\ref{sec: theorem_aug}), instruction-following retrieval (\S\ref{sec:if}), and traditional text embedding tasks (\S\ref{sec:emb}).

\subsection{Long-document Retrieval}
\label{sec:long}

Encoding long documents is crucial for retrieval, yet prior embedding models are typically limited to inputs of 512 tokens \citep{longembed}. To adapt diffusion LMs and LLMs for long-document retrieval, we train the \ours and LLM embedding models on a subset\footnote{Simulating low-resource settings, we train on the first N samples, which shows to capture most of performance.} of the Public E5 dataset \citep{echo}, using an input length of 4,096 tokens. This dataset is designed for developing general-purpose text embedding models and spans a wide range of tasks and document lengths (\Cref{training_sets}). We evaluate embedding models on the \longembed benchmark \citet{longembed}, which includes two synthetic tasks (\ie finding the document containing a personalized passkey or fact needle) and four real-world tasks (\ie finding the document containing a correct answer or summarization), featuring documents of varying lengths and dispersed target information.

\begin{figure*}[t!]
    \centering
    \includegraphics[width=1\textwidth]{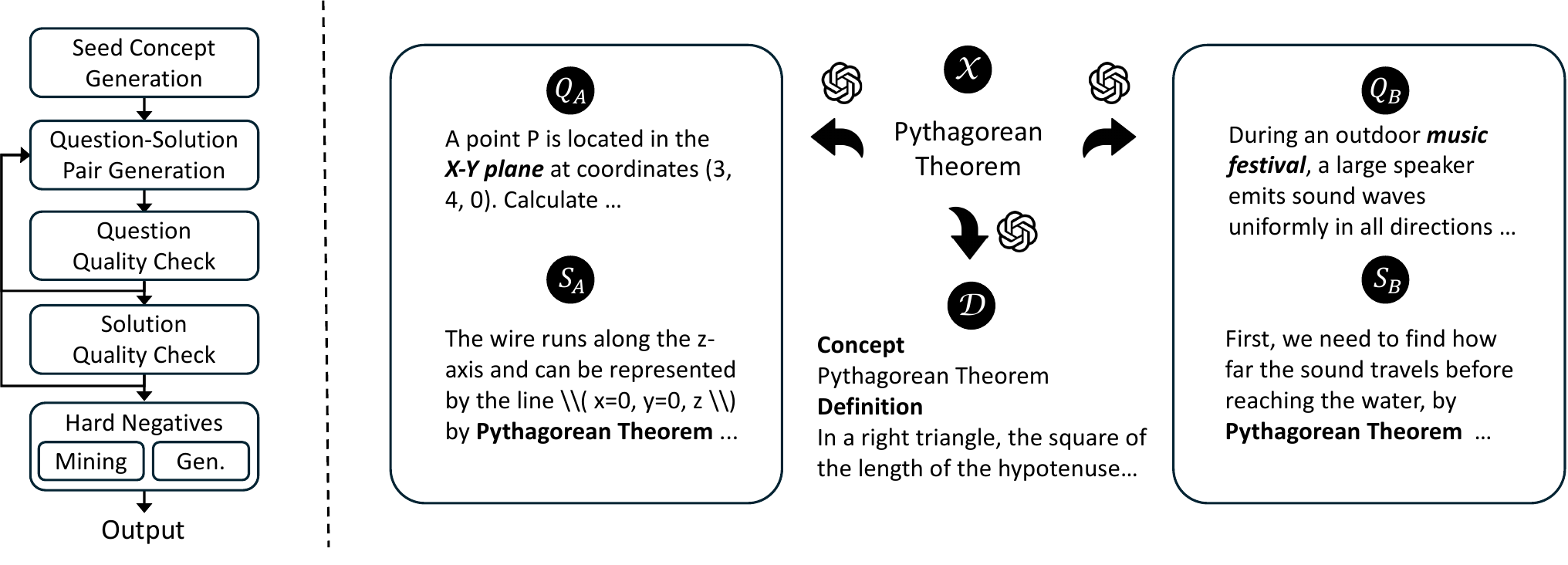}
    \captionsetup{justification=justified, singlelinecheck=false}  %
    \caption{\textbf{Left}: data augmentation pipeline. \textbf{Right}: qualitative examples of seed concepts, their definitions, and associated question–solution pairs. A question-to-question retrieval sample can be constructed using a query question $Q_A$, a positive document $(Q_B, S_B)$ generated from the same concept $X$, and a hard negative document $(Q_C, S_C)$ generated from a different concept $X'$. A question-to-concept retrieval sample can consist of a query question $Q_B$, a positive document $D$ (the definition of the relevant concept), and a hard negative document $D'$ (the definition of a different concept $X'$).}
    \label{fig:pipeline}
\end{figure*}

\subsection{Reasoning-intensive Retrieval}
\label{sec: theorem_aug}

Many real-world search queries require in-depth reasoning beyond lexical or semantic matching. \citet{bright} introduces the \bright benchmark for this type of reasoning-intensive retrieval, \eg finding a tutorial for a math question related to the same theorem (TheoQ.) or reference code that implements the same algorithm as the programming problem (Leet.). However, at the time of this work, no existing dataset effectively supports this task.\footnote{ReasonIR \cite{reasonir} recently releases a dataset built on the \bright corpus, which we will use for further evaluation. In contrast, we do not use any \bright document.} Thus, we develop a new dataset \theoremaug using LLMs\footnote{We experimented with GPT-4o-mini and DeepSeek-V3 for query generation. GPT-4o-mini produced more diverse and complex questions, so it was used for the final dataset.} for training embedding models for logical reasoning. \theoremaug mainly contains two types of task samples: (1) question-to-concept: retrieving the concept that is helpful for solving the given question; (2) question-to-question: retrieving the question that can be solved using the same concept as the given question. The trained models are evaluated on the math and code subsets of \bright (\eg TheoremQA and Leetcode). The dataset construction includes following steps:

\paragraph{Concept, Question, and Solution Generation.} As shown in \Cref{fig:pipeline} Left, we begin by prompting LLMs to generate a list of concepts such as mathematical and physics theorems, programming algorithms, and financial equations (see \Cref{all_theorems}). For each seed concept, we create one definition and eight question-answer pairs as shown in \Cref{fig:pipeline} Right, with the option of generating more. Our prompts are carefully crafted to produce questions that are novel, challenging, and varied in type (theoretical or applied), background context, and length—while remaining focused on the seed concept. In question-to-question retrieval, each question is used as a query, with other questions from the same concept serving as positives. For question-to-concept retrieval, the corresponding definition is used as the positive document. Finally, our data leakage analysis (see \Cref{leak}) confirms that there are no overlapping questions with \bright.

\paragraph{Question and Solution Quality Check.} We conduct a quality examination with two primary objectives: (1) Relevance — the generated question and solution should involve the seed concept; and (2) Correctness — the solution should correctly address the generated question. LLMs are instructed to discard any irrelevant or incorrect generations and produce appropriate replacements.

\paragraph{Hard Negative Mining and Generation.} Hard negative documents are crucial for promoting fine-grained relevance discrimination. We construct hard negatives through both document mining and LLM generation. For mining, we retrieve lexically similar questions from the entire generation set using BM25 \citep{bm25}, ensuring they are based on irrelevant and diverse concepts. For generation, we prompt LLMs to create novel questions with similar background but unrelated to the seed concept, ensuring they are not helpful for answering the query. 

In total, we curate 10,896 triplets (query, positive, hard negative), with statistics in \Cref{training_sets} and examples in \Cref{theorem_ex}. The data creation prompts are described in \Cref{aug_prompts}.

\begin{table*}[t!]
\centering
\renewcommand{\arraystretch}{1.1}
\captionsetup{justification=justified,singlelinecheck=false}
\begin{tabularx}{\textwidth}{
    >{\raggedright\arraybackslash}l 
    >{\centering\arraybackslash}c 
    *{4}{>{\centering\arraybackslash}X} 
    *{4}{>{\centering\arraybackslash}X} 
    >{\centering\arraybackslash}c
} 
\toprule
\multirow{2}{*}{\textbf{}} & \multirow{2}{*}{\textbf{\textit{N}}} & \multicolumn{4}{c}{\textbf{Synthetic (Acc@1)}} & \multicolumn{4}{c}{\textbf{Real (nDCG@10)}} & \multirow{2}{*}{\textbf{Avg.}} \\
\cmidrule(lr){3-6} \cmidrule(lr){7-10}
                           &                                      & \textbf{Pk$\leq$4k} & \textbf{Pk>4k} & \textbf{Nd$\leq$4k} & \textbf{Nd>4k} & \textbf{NQA} & \textbf{QMS} & \textbf{SFD} & \textbf{WQA} &                         \\
\hline
E5-Mistral    & 2M      &   95.6 & 30.0 &   68.8 & 14.0 & 44.6 & 43.6 & 96.8 & 82.0 &   59.4    \\
\hline
Llama3     &  16k &   54.0  & 3.3 & 51.2   &    4.0  & 31.9 & 30.8  & 67.6 & 47.3 &  32.3       \\        
 \; + LLM2Vec    & 16k  & 59.6 & 4.0 &    59.2 & 6.7   & 36.5 & 34.4 & 84.5 & 51.1 & 42.0   \\
Mistral &  16k & 87.6 & 22.0 & 72.4 & 20.0 & 35.0 & 36.8 & 85.4 & 62.9 & 52.8 \\
\makecell[l]{ \; + LLM2Vec} & 16k  & 98.8 & \textbf{30.0} & 69.2 & \textbf{21.3} & 39.5 & 40.0 & 91.6 & \textbf{78.0}  & 58.6 \\
Qwen2.5 & 16k & 86.8 & 24.6 & 68.0 & 19.3 & 31.7 & 35.6 & 88.0 & 68.3 &  52.8 \\
 \; + LLM2Vec   & 16k & 94.0  & 28.7   &  72.4  & 20.0  & 36.5 & 37.3 & 90.5 & 68.1 &  55.9   \\
 \rowcolor{gray!20}
\ours         &   16k    &  \textbf{100} &  29.3 & \textbf{86.8}  & 20.0   &  \textbf{42.1} & \textbf{43.8} & \textbf{98.0} & 77.2 &   \textbf{62.2}    \\
\bottomrule
\end{tabularx}
\caption{Results on long document retrieval (\longembed) for embedding models trained with \textbf{E5} text embedding data. \textit{N} is the amount of E5 samples used for training. \textit{Pk$\leq$4k} refers to \textit{Passkey} Retrieval with an evaluation length less than or equal to 4,096. \textit{Nd>4k} refers to \textit{Needle-in-a-haystack} Retrieval with length larger than 4,096. \textit{NQA}, \textit{QMS}, \textit{SFD}, \textit{WQA} is short for \textit{NarrativeQA}, \textit{QMSum}, \textit{SummScreenFD}, \textit{2WikiMultihopQA}, respectively. The best results with 16k training data are in \textbf{bold}.}
\label{longembed_results}
\end{table*}

\subsection{Instruction-Following Retrieval}
\label{sec:if}

Inspired by progress in instruction-tuned LMs, there is a growing demand for instruction-following retrievers. To this end, \citet{promptriever} develop a instance-level instruction training set MS MARCO with Instructions. We train models using a subset of MS MARCO with Instructions and evaluate their instruction-following capabilities on the \followir \citep{followir} benchmark.

\subsection{Traditional Text Embedding Tasks}
\label{sec:emb}

In addition to retrieval, we compare diffusion and LLM embedding models on the traditional text embedding tasks. Following \citet{llm2vec}, we train the models using the same subset of the Public E5 dataset as in \Cref{sec:long}, and evaluate them on the Massive Text Embedding Benchmark (MTEB) \citep{mteb}, which includes tasks such as reranking, clustering, classification, semantic textual similarity, and more.

\section{Experiments}
\label{sec:exp}

As detailed in \Cref{setups}, we compare \ours and LLM embeddings in four setups for different capabilities. In each setup, we fine-tune the models using one dedicated dataset.

\subsection{Models}
We take \dream-v0-Instruct-7B \citep{dream} as the backbone model for \ours. Initialized from Qwen2.5-7B \citep{qwen}, \dream is pre-trained on 580 billion tokens\footnote{Considering this corpus is significantly smaller than the multi-trillion-token pre-training of Qwen2.5, we believe this phase primarily serves to adapt the generation paradigm from autoregressive to diffusion, rather than to introduce substantial new knowledge. It does not overlap with benchmark data.} covering text, code, and math, and achieves performance competitive with leading AR LMs of similar scale. For comparison, we include similarly sized autoregressive models: Llama-3-8B-Instruct \citep{llama3}, Mistral-7B-Instruct-v0.2 \citep{mistral}, and Qwen2.5-7B-Instruct \citep{qwen}. 
We also include LLM2Vec \citep{llm2vec}, which adapts LLMs to use bidirectional attention via two intermediate pre-training steps: Masked Next Token Prediction (MNTP) and unsupervised contrastive learning \citep{simcse} using Wikitext data. We focus on MNTP, which has greater impact on final performance in line with \citet{llm2vec}, and report unsupervised learning results in \Cref{sec: simcse}. The evaluation metrics and implementation details are elaborated in \Cref{implementation}.

\subsection{Long Document Retrieval}

\Cref{longembed_results} compares \ours and LLM embedding models on the \longembed benchmark. We find that LLM2Vec brings an average 6.2\% gain for LLM embedding models, this is because learning to use bidirectional attention introduces potential benefits such as capturing long-range dependencies.
Our synthetic experiments further reveal that, although models are trained with an input length of 4,096 tokens, LLM embedding models may still fail to encode key information effectively \citep{needle}. As expected, \ours with bidirectional attention during large-scale pretraining, ace the synthetic tasks (with 100\% and 86.8\% accuracy) and outperforms all LLM embedding models on real-world tasks. %

\begin{table*}[t!]
\centering
\renewcommand{\arraystretch}{1.1}
\captionsetup{justification=justified,singlelinecheck=false}
\begin{tabularx}{\textwidth}{
    >{\raggedright\arraybackslash}l 
    >{\centering\arraybackslash}X 
    >{\centering\arraybackslash}X 
    >{\centering\arraybackslash}X
    >{\centering\arraybackslash}X
    >{\centering\arraybackslash}X
    >{\centering\arraybackslash}X
    >{\centering\arraybackslash}X
} 
\toprule
 &\textbf{\textit{N}} & \textbf{TheoT.} & \textbf{TheoQ.} & \textbf{AoPS}  & \textbf{Leet.} & {\textbf{Average}}    \\
\hline
E5-Mistral   &  \multirow{2}{*}{-}   &    26.8 & 26.1 & 7.1 & 28.7    &  22.2  \\
ReasonIR   &   &    27.2 & 31.9 & 14.7 & 35.0    &  27.2 \\
\hline
Llama3     &   10k  &   31.0      &  34.6   & 5.8 &  18.9  & 22.6\\
\makecell[l]{\; + LLM2Vec}     & 10k & 28.3   &    33.8 &  12.6  &  23.6 & 24.6  \\
Mistral & 10k & 32.4 & 33.7 & 9.4 & 20.7 & 24.1 \\
\makecell[l]{\; + LLM2Vec}  & 10k  & 30.8 & 30.4 &  9.1 & 22.9 & 23.3  \\
Qwen2.5      &   10k  &   34.7   &   40.2    & \textbf{15.5} &  \textbf{32.0} & 30.6 \\
\makecell[l]{ \; + LLM2Vec} & 10k  &   32.3 &  37.2 &  10.3 &   31.0 & 27.7  \\
 \rowcolor{gray!20}

\ours     & 10k   & \textbf{38.9}  & \textbf{48.3} & 15.4 & 30.0  & \textbf{33.2} \\
\bottomrule
\end{tabularx}
\caption{Results on theorem-based benchmarks in reasoning-intensive retrieval (\bright) for embedding models trained using \theoremaug. We report nDCG@10 for \textit{TheoremQA} with with theorem retrieval (\textit{TheoT.}) and question retrieval (\textit{TheoQ.}), \textit{AoPS}, and \textit{LeetCode} (\textit{Leet.}). \textit{N} is the amount of \theoremaug samples used for training. The best results with 10k training data are in \textbf{bold}.}
\label{bright_results}
\end{table*}

\begin{table*}[t!]
\centering
\renewcommand{\arraystretch}{1.1}
\captionsetup{justification=justified,singlelinecheck=false}
\begin{tabularx}{\textwidth}{
    >{\raggedright\arraybackslash}l 
    >{\centering\arraybackslash}c 
    *{2}{>{\centering\arraybackslash}X} 
    *{2}{>{\centering\arraybackslash}X} 
    *{2}{>{\centering\arraybackslash}X} 
    *{2}{>{\centering\arraybackslash}X}
} 
\toprule
\multirow{2}{*}{\textbf{}} & \multirow{2}{*}{\textbf{\textit{N}}} & \multicolumn{2}{c}{\textbf{Robust04}} & \multicolumn{2}{c}{\textbf{News21}} & \multicolumn{2}{c}{\textbf{Core17}} & \multicolumn{2}{c}{\textbf{Average}} \\
\cmidrule(lr){3-4} \cmidrule(lr){5-6} \cmidrule(lr){7-8} \cmidrule(lr){9-10}
      &  &      MAP  & MRR &  nDCG  & MRR  &  MAP  & MRR &     Score  & MRR      \\
\hline
\makecell[l]{Prompt\\-riever}  & 1M   &  28.3   &  +11.7    &  28.5  & +6.4   & 21.6 & +15.4  & 26.1   & +11.2   \\
\hline
Llama3   &   16k   & 12.7 & +2.3 & 17.6 & -1.0 & 11.7 & +1.7 & 14.0 & +1.0 \\
\makecell[l]{\; + LLM2Vec}   & 16k & 17.6 & +1.9 & 21.6 & +1.2 & 16.7 & +4.7 & 18.6 & +2.6  
\\
Mistral & 16k & \textbf{21.4} & \textbf{+7.0} & 27.4 & +1.1 & 16.0 & +7.2 & \textbf{21.5} & \textbf{+5.1} 
\\
\makecell[l]{\; + LLM2Vec}  & 16k & 20.7 & +5.7 & 25.1 & +1.1 & \textbf{16.3} & \textbf{+7.5} &  20.7 & +4.8 \\
Qwen2.5  & 16k & 15.6 & +1.4 & 20.4 & +1.7 & 13.9 & +4.4 & 16.6 & +2.5 
\\
\makecell[l]{\; + LLM2Vec}  & 16k & 14.5 & +1.6 & 19.7 & +0.1 & 13.8 & +2.8 & 16.0 &  +1.5  \\
 \rowcolor{gray!20}
\ours         & 16k  & 18.9 & +5.7 & \textbf{27.7} & \textbf{+3.6} & 16.2 & +6.0  & 20.9 & \textbf{+5.1} \\
\bottomrule
\end{tabularx}
\caption{Results on instruction-following document retrieval (\followir) for embedding models trained using \textbf{\msmarco with Instructions}. \textit{N} is the amount of \msmarco samples used for training. MAP@1000/nDCG@5 range from 0-100. MRR is short for the pairwise evaluation metric measuring instruction following when instructions change, ranging from -100 to 100 (higher is better). The best results trained with 16k data are in \textbf{bold}.}
\label{if_results}
\end{table*}

\subsection{Reasoning-Intensive Retrieval}
\label{sec: reasoning}

Compared to \longembed, the logical reasoning-based tasks\footnote{Other \bright tasks, such as StackExchange (non-logical content) and Pony (a niche programming language), differ significantly from the domain of \theoremaug, thus we have excluded them. Our results show that there is no significant performance gains on these tasks using \theoremaug.} in \bright involve shorter queries but more complex documents, require stronger reasoning capabilities. According to \Cref{bright_results}, LLM2Vec is ineffective, likely due to limited data scale and complexity in its continue pre-training. All fine-tuned embedding models outperform the general-purpose E5-Mistral and perform on par with the reasoning-focused retriever ReasonIR \citep{reasonir}, highlighting the effectiveness of our lightweight dataset \theoremaug. Among LLM embeddings, Qwen2.5 achieves the highest overall performance, likely benefiting from its strong mathematical and coding capabilities. \ours further improves on Qwen2.5, with gains of +4.2 and +8.1 points on the TheoT. and TheoQ. tasks, substantially outperforming the rest models. These improvements suggest that the ability to attend to both past and future context is essential for understanding complex logic, theorems and equations. On AoPS and Leet., \ours shows relatively smaller improvements, as discussed further in \Cref{analysis} Q3.

\subsection{Instruction-Following Retrieval}
\label{result: if}

As shown in \Cref{if_results}, Mistral achieves the best performance among the models. As noted by \citet{llm2vec}, bidirectional attention may be effective for Mistral even without LLM2Vec pre-training. \ours demonstrates comparable instruction-following ability to Mistral (\ie +5.1 pairwise MRR). We hypothesize that the limited length\footnote{Following Promptriever, we use inputs of max 304 tokens.} and low complexity of queries and documents constrain the benefits of \ours's bidirectional embedding approach.

\subsection{General Text Embedding Tasks}

We evaluate the \ours and LLM embedding models on 15 general text embedding tasks on MTEB. According to \Cref{general_text_embed}, \ours performs on par with LLM embedding models, as expected. Similar to \Cref{result: if}, most traditional tasks involve shorter inputs and less reasoning, limiting \ours's benefits.

\section{Analysis}
\label{analysis}

\paragraph{Q1: Why do generalist embedding models underperform in reasoning-intensive retrieval?}
Although identifying questions that share the same theorem resembles a clustering task, traditional embedding models are not trained to recognize theorems or use them as clustering signals. As shown in \Cref{fig:t-sne}, the E5-Mistral embeddings of \theoremaug documents are more dispersed, lacking clear cluster boundaries\footnote{Most \theoremaug questions focus on a single concept, which is generally distinct from others; therefore, clear boundaries among concept clusters are expected.}. Furthermore, the case study in \Cref{case_1} shows that E5-Mistral often relies on superficial lexical cues (e.g., exact numbers like ``10'' or keywords like ``smallest'') and shallow semantic patterns (e.g., ``what is the count'' vs. ``how many'') when matching questions. Detailed studies are presented in \Cref{case_study}.

\paragraph{Q2: How important is bidirectional attention for different models and tasks?} 
To assess the role of bidirectional attention, we ablate two embedding models, Mistral and \ours on \bright. For each model, we compare testing performance using full bidirectional (\overcurvedleftarrow{\text{causal}} and \undercurvedrightarrow{\text{reverse}}) attention versus unidirectional (causal only) attention. According to  \Cref{atten_ablation}, one notable finding is that \ours exhibits a substantially larger performance drop when reverse attention is disabled, which suggests that \ours depends more heavily on bidirectional context, likely due to its bidirectional pre-training.
Furthermore, tasks differ in their sensitivity to reverse attention. Performance on the Leet. task remains relatively stable (even increase) without reverse attention, while performance on the TheoQ. task degrades significantly. This indicates the bidirectional attention is critical for logical reasoning tasks like TheoremQA, supporting the performance gains observed for \ours on TheoT. and TheoQ. in \Cref{sec: reasoning}.

\begin{table}[t!]
\centering
\renewcommand{\arraystretch}{1.1}
\captionsetup{justification=justified,singlelinecheck=false}
\begin{tabularx}{0.48\textwidth}{
    >{\raggedright\arraybackslash}l 
    >{\centering\arraybackslash}X
    >{\centering\arraybackslash}X
} 
\toprule
 & \textbf{Bidirectional} & \textbf{Unidirectional} \\
\hline
\multicolumn{3}{c}{Mistral} \\
\textbf{TheoT.} & 32.4  & 4.0 (-28.4) \\
\textbf{TheoQ.} & 33.7 & 9.6 (-24.1) \\
\textbf{AoPS} & 9.4 & 7.8 (-1.6)  \\
\textbf{Leet.}& 20.7 & 34.9 (+14.2) \\
\hline
\multicolumn{3}{c}{\textit{\ours}} \\
\textbf{TheoT.}& 38.9  & 1.1 (-37.8) \\
\textbf{TheoQ.}& 48.3  & 0.7 (-47.6) \\
\textbf{AoPS} & 15.4 & 2.8 (-12.6) \\
\textbf{Leet.}& 30.0 & 29.5 (-0.5) \\
\bottomrule
\end{tabularx}
\caption{Ablation study comparing \bright performance when using bidirectional vs. unidirectional attention at test time. Both models are trained with bidirectional attention and \theoremaug data.}
\label{atten_ablation}
\end{table}

\paragraph{Q3: Why are improvements less evident in Leet. and AoPS compared to TheoQ. in \bright?} 
To better understand the difference between TheoQ., Leet. and AoPS, we conduct a comparative analysis in \Cref{count_comparison} and a human re-evaluation for gold documents in \Cref{human_reeval}. Our analysis reveals notable noise in the gold annotations and corpora of Leet. and AoPS, impacting the evaluation results (see \Cref{noise_data}). In particular, we observe multiple types of relevance (\eg contextual and algorithmic) annotated in Leet., while it is more consistent in TheoT. and TheoQ..

\paragraph{Q4: Does the performance gain diminish as the training dataset size increases?} 
\Cref{trend} illustrates the TheoQ. performance of \ours and its base model, Qwen2.5, as the training size varies from 2k to 10k. The performance gap remains substantial within this range, suggesting the LLM-based models may need extensive data to learn effective bidirectional attention as \ours.

\section{Conclusion}

This paper presents the first exploration of using diffusion language models as text embedding models—a intuitively motivated direction. Through extensive experiments on long-document retrieval (\longembed), reasoning-intensive retrieval (\bright), instruction-following retrieval (\followir), and general embedding tasks (MTEB), we demonstrate the advantages of diffusion embeddings over LLM embeddings in capturing global context for long and complex text. We attribute these gains to large-scale bidirectional pre-training. We hope this work provides meaningful insights for both the text embedding community and the development of diffusion LMs.

\section*{Limitations}

We evaluate only the state-of-the-art diffusion LM, \dream \citep{dream}. Other models, such as LLaDA \citep{llada}, are expected to exhibit inferior text embedding performance given their comparatively weaker generative and reasoning abilities. Due to resource constraints, we limit the training scale within 20k samples. Larger-scale experiments with millions of examples could reveal further insights. In \theoremaug, irrelevant or incorrect documents may still remain when LLMs fail to identify them during the quality check steps. However, contrastive training is resilient to a certain degree of noise in the data.

\bibliography{main.bbl}

\clearpage
\appendix
\onecolumn

\begin{table}[h]
\centering
\renewcommand{\arraystretch}{1.1}
\captionsetup{justification=justified,singlelinecheck=false}
\begin{tabularx}{0.95\textwidth}{
    >{\raggedright\arraybackslash}l 
    >{\centering\arraybackslash}X
    >{\centering\arraybackslash}X
    >{\centering\arraybackslash}X
} 
\toprule
\textbf{Task} & \textbf{Llama3} & \textbf{Mistral} & \textbf{Dream} \\
\hline
ArguAna & 57.33 & 54.46 & 58.47 \\
Banking77Classification & 87.35 & 87.03 & 87.50\\
BiorxivClusteringS2S & 38.45 & 35.80 & 35.56 \\
EmotionClassification & 47.81 & 50.81 & 51.44 \\
MassiveIntentClassification & 78.69 & 76.89 & 75.64 \\
MedrxivClusteringS2S & 33.46 & 31.26 & 31.42 \\
NFCorpus & 38.04 & 39.27 & 35.44 \\
SciDocsRR & 85.19 & 84.41 & 82.00 \\
SciFact & 75.93 & 71.21 & 71.42 \\
SICK-R & 80.64 & 78.92 & 76.49 \\
SprintDuplicateQuestions & 92.30 & 95.28 & 95.30 \\
StackOverflowDupQuestions & 55.32 & 54.32 & 50.69 \\
STS17 & 87.30 & 88.15 & 87.83 \\
STSBenchmark & 84.76 & 85.47 & 83.43 \\
TwentyNewsgroupsClustering & 53.28 & 49.50 & 52.56 \\
\hline
Average & 66.39 & 65.52 & 65.01 \\
\bottomrule
\end{tabularx}
\caption{Evaluation on 15 general text embedding tasks selected by \citet{llm2vec} from MTEB for Llama3 + LLM2Vec, Mistral + LLM2Vec, and Dream models, which are trained with 16k Public E5 data and bidirectional attention.}
\label{general_text_embed}
\end{table}

\vspace{1cm}

\begin{figure}[h]
    \centering
    \includegraphics[width=0.5\textwidth]{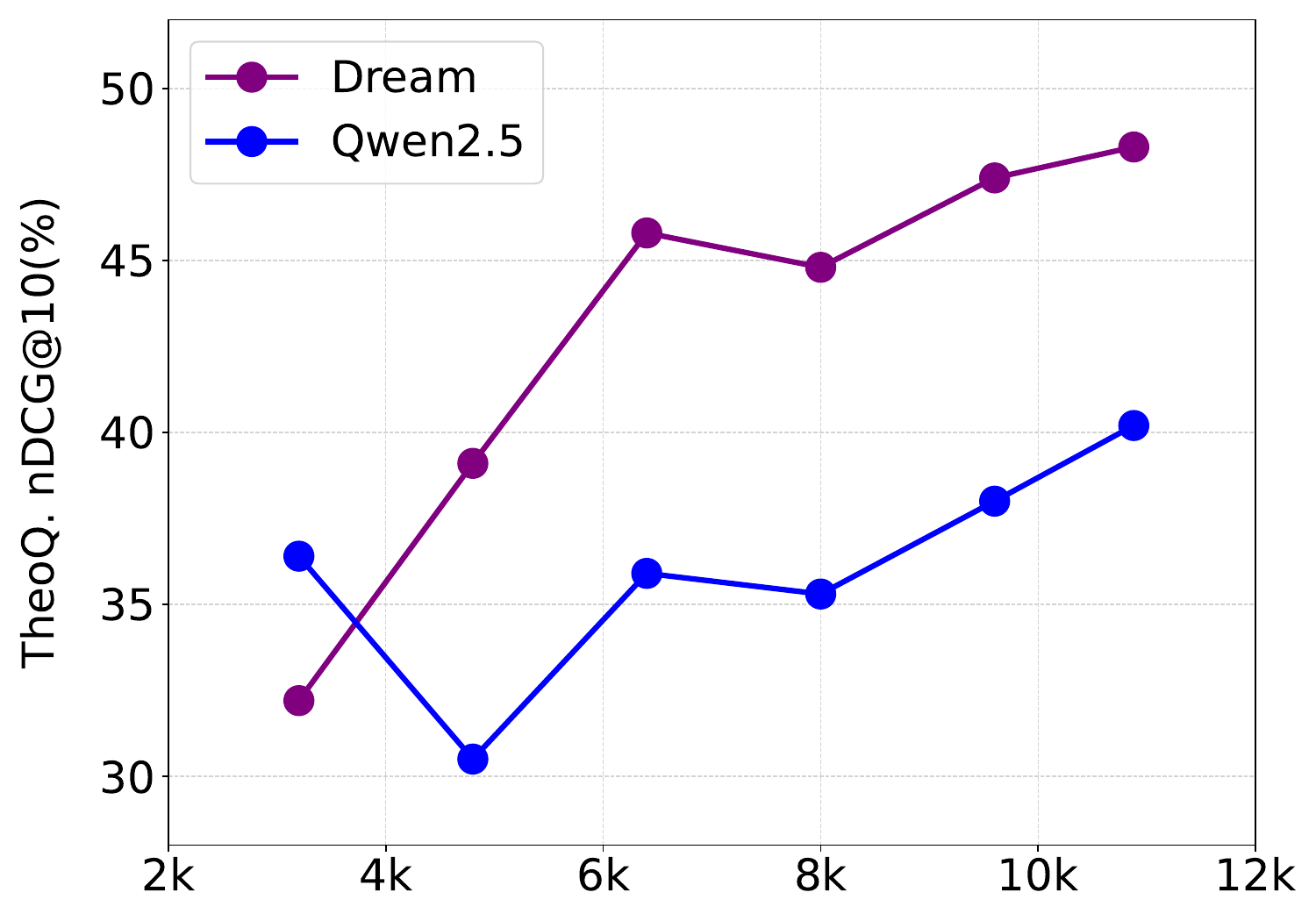}
    \captionsetup{justification=justified, singlelinecheck=false}  %
    \caption{Retrieval performance on TheoQ. for Dream and Qwen2.5 models trained with varying amounts of \theoremaug data.}
    \label{trend}
\end{figure}

\twocolumn

\section{\theoremaug Seed Concept List}
\label{all_theorems}

\paragraph{60 Algorithms:} 
    Sweep Line Algorithm,
    Kahn's Algorithm,
    Dijkstra's Algorithm,
    Game Theory,
    Two Pointers,
    N-Queens Problem,
    Depth First Search (DFS),
    Prefix Sum,
    Greedy Algorithm,
    Bucket Sort,
    Breadth First Search (BFS),
    Longest Common Subsequence (LCS),
    Huffman Coding,
    Manhattan Distance,
    Topological Sorting,
    Rod Cutting Problem,
    Binary Search,
    Knapsack Algorithm (0/1 Knapsack),
    Floyd-Warshall Algorithm,
    Bellman-Ford Algorithm,
    Merge Sort,
    Quick Sort,
    Heap Sort,
    Bubble Sort,
    Insertion Sort,
    Selection Sort,
    Kruskal's Algorithm,
    Prim's Algorithm,
    Kadane's Algorithm,
    Rabin-Karp Algorithm,
    Knuth-Morris-Pratt (KMP) Algorithm,
    Boyer-Moore Algorithm,
    Longest Increasing Subsequence (LIS),
    Edit Distance,
    Sieve of Eratosthenes,
    Tarjan's Algorithm,
    Kosaraju's Algorithm,
    Z Algorithm,
    LRU Cache Algorithm,
    A-star search algorithm,
    Hamiltonian Path,
    Substring Search Algorithm,
    Permutations,
    Combinations,
    Knapsack DP with Bitmasking,
    Matrix Exponentiation,
    Square Root Decomposition,
    Mo's Algorithm,
    Strassen's Algorithm,
    K-Means Clustering,
    Gradient Descent,
    Support Vector Machines (SVM),
    Aho-Corasick Algorithm,
    Ford-Fulkerson Algorithm,
    Trapping Rain Water,
    Matrix Chain Multiplication,
    Coin Change Problem,
    Palindrome Partitioning,
    Sudoku Solver,
    Newton's Method.

\begin{figure}[t!]
    \centering
    \begin{subfigure}{0.48\textwidth}
        \includegraphics[width=\linewidth]{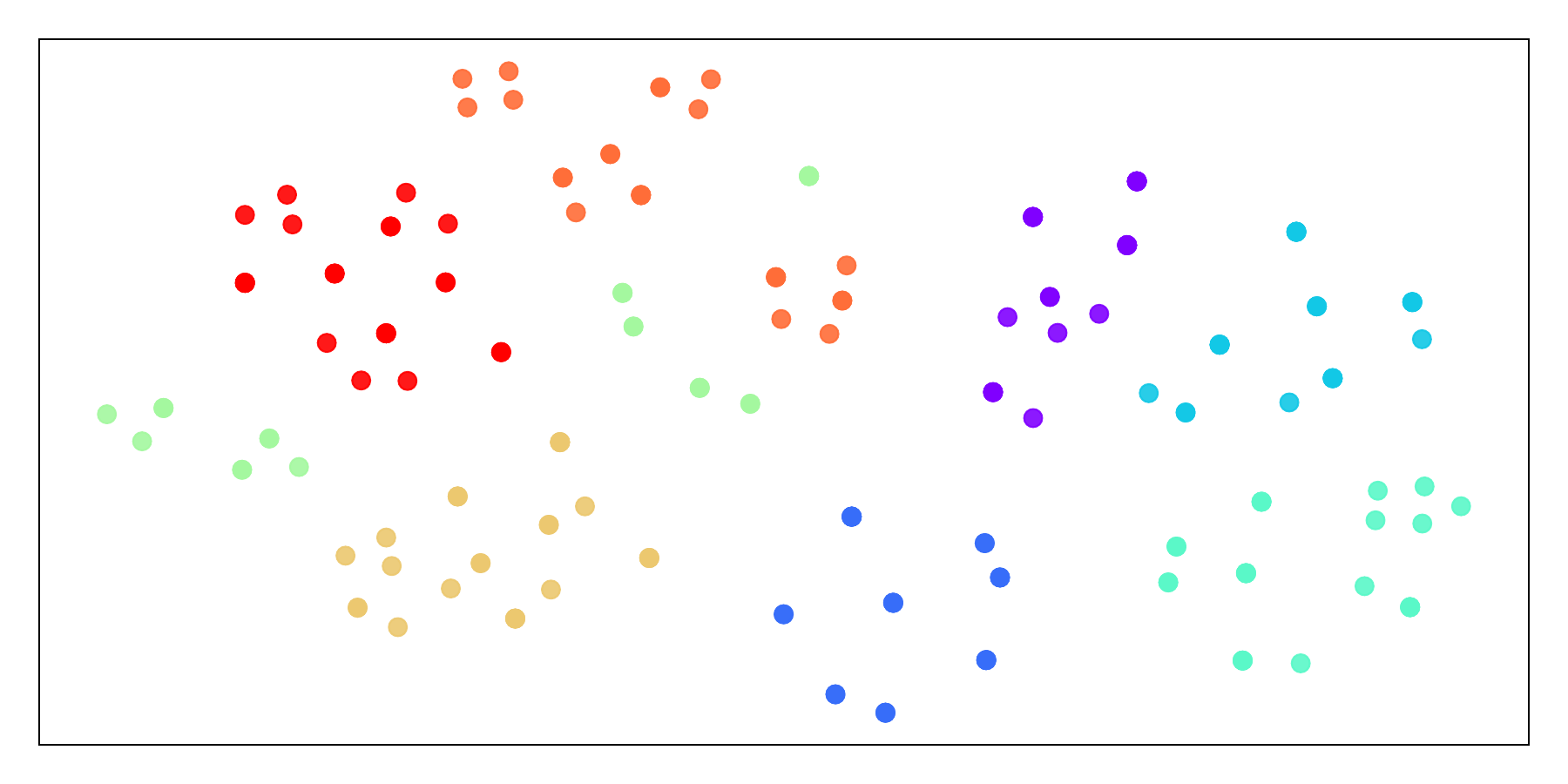}
        \caption{E5-Mistral without \theoremaug fine-tuning.}
        \label{e5_embed}
    \end{subfigure}

    \vspace{0.5em} %

    \begin{subfigure}{0.48\textwidth}
        \includegraphics[width=\linewidth]{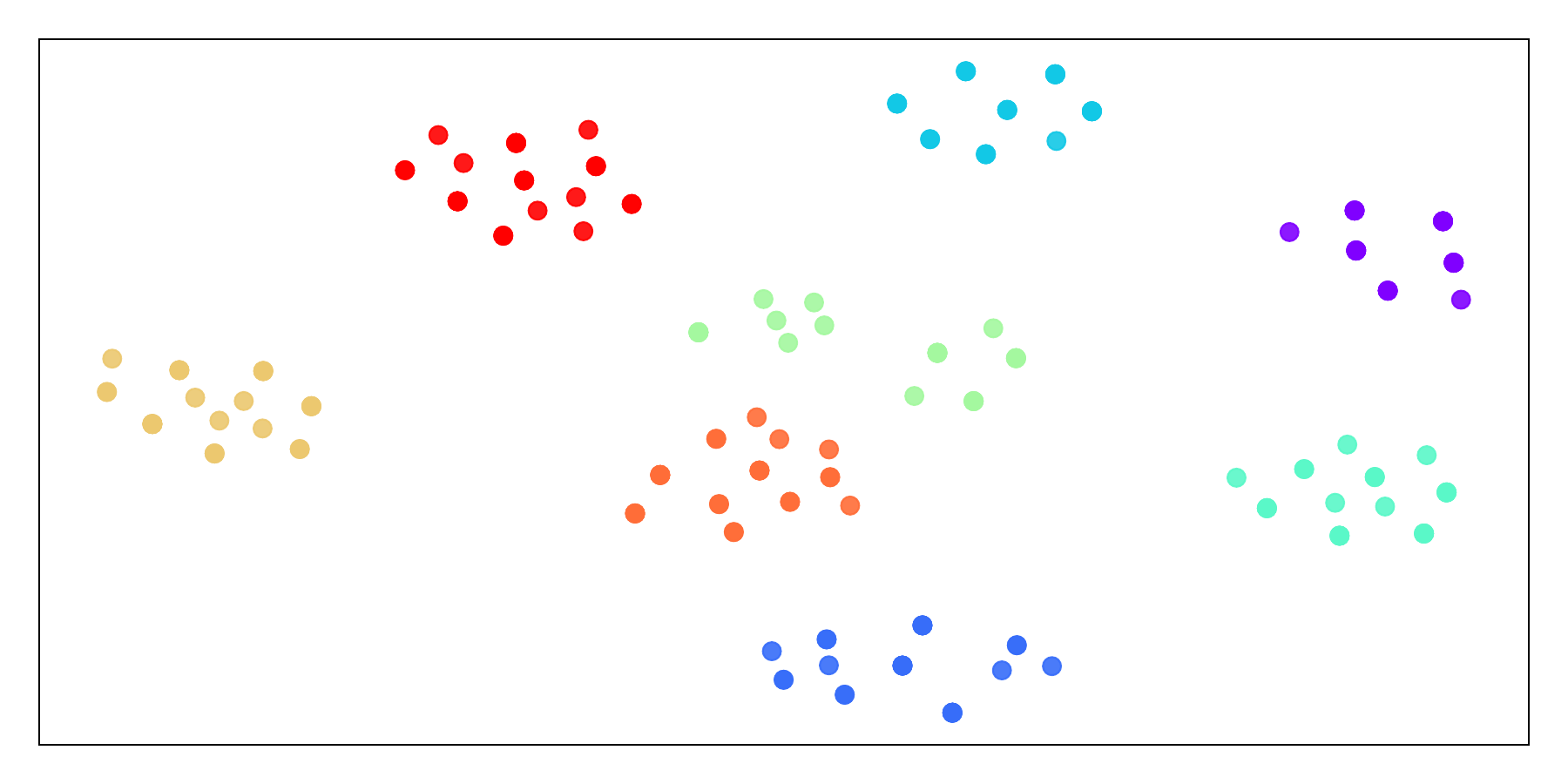}
        \caption{\ours fine-tuned on \theoremaug.}
    \end{subfigure}

    \caption{t-SNE visualization of document embeddings from \theoremaug. The documents are grouped and color-coded by concept (see legend in \Cref{color_mapping}). Mathematical theorems include \textit{Vieta’s Formulas}, \textit{Pigeonhole Principle}, \textit{Euler’s Identity}, and \textit{Central Limit Theorem}. Algorithmic concepts include \textit{Two Pointers}, \textit{N-Queens Problem}, \textit{Sweep Line Algorithm}, and \textit{Kahn’s Algorithm}.}

    \label{fig:t-sne}
\end{figure}

\begin{figure}[t!]
    \centering
    \includegraphics[width=0.45\textwidth]{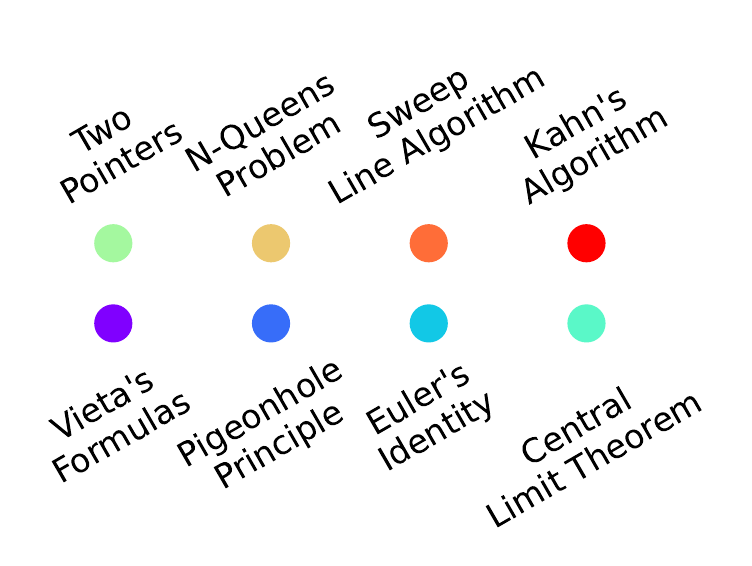}
    \captionsetup{justification=justified, singlelinecheck=false}  %
    \caption{The concept color mapping for \Cref{fig:t-sne}.}
    \label{color_mapping}
\end{figure}

\paragraph{90 Math Theorems:} Newton's Sums,
    Pigeonhole Principle,
    Chicken McNugget Theorem,
    Simon's Favorite Factoring Trick,
    Fermat's Little Theorem,
    Ptolemys theorem,
    Euler's Identity,
    Euclidean algorithm,
    Cauchy-Riemann Equations (Complex Analysis),
    Vieta's Formulas,
    Triangle Inequality,
    Power of a Point,
    Central Limit Theorem,
    Pick's Theorem,
    Shoelace Theorem,
    Legendre's formula,
    Principle of Inclusion Exclusion,
    Ceva's Theorem,
    Logarithm: Change of Base Formula,
    Stars and Bars formula,
    Eigenvalue equation,
    Intermediate Value Theorem,
    Mass point geometry theorem,
    Geometric probability in 2D,
    Fourier Transform,
    Cramer's Rule,
    Vertex cover in graph theory,
    One-sample t-test,
    Z-transform,
    Ramsey's Theorem,
    Pollard's Rho Algorithm (Factorization),
    Chinese Remainder Theorem,
    Taylor's Theorem,
    Addition of Multiindices,
    Bayes' Theorem,
    Binomial Theorem,
    Mean Value Theorem for Derivatives,
    Pythagorean Theorem,
    Lagrange's Theorem (Group Theory),
    The Chain Rule in calculus,
    Green's Theorem,
    Cauchy-Schwarz Inequality,
    Divergence Theorem,
    Second Part of the Fundamental Theorem of Calculus (FTC2),
    Quadratic Formula (for polynomials of degree 2),
    Fundamental Theorem of Arithmetic,
    Rolle's Theorem,
    De Moivre's Theorem,
    Law of Large Numbers,
    Cayley-Hamilton Theorem,
    L'Hôpital's Rule,
    Singular Value Decomposition (SVD) Theorem,
    The Squeeze Theorem,
    Brouwer Fixed-Point Theorem,
    Tychonoff's Theorem,
    Bézout's Theorem,
    Vandermonde's Identity,
    Wilson's Theorem,
    Markov Property,
    Invertible Matrix Theorem,
    Sylow Theorems,
    Cantor's Theorem,
    Heron's Formula (for the area of a triangle),
    Laplace Transform,
    Bolzano-Weierstrass Theorem,
    Weierstrass Approximation Theorem,
    Cauchy's Mean Value Theorem,
    Lindelof's Theorem,
    Poisson Limit Theorem,
    Mertens' Theorem,
    Chebyshev's Inequality,
    Markov's Inequality,
    Jensen's Inequality,
    Borel-Cantelli Lemma,
    Chauvenet's Criterion,
    Helly's Theorem,
    Holder's Inequality,
    Minkowski's Inequality,
    Euler's Formula for Planar Graphs,
    Hahn Decomposition Theorem,
    Radon-Nikodym Theorem,
    Kakutani Fixed-Point Theorem,
    Sum of a Geometric Series Formula,
    The Isoperimetric Inequality,
    Spectral Theorem,
    Power Rule (for derivatives),
    Hadamard's Determinant Theorem,
    Borel's Theorem,
    Runge's Theorem,
    Euler's Formula for Polyhedra,
    Integral of a Power Function Formula,
    Poincaré Recurrence Theorem,
    Extreme Value Theorem,
    Dirichlet's Theorem on Primes in Arithmetic Progressions,
    Lefschetz Fixed-Point Theorem,
    Seifert-van Kampen Theorem,
    Hurewicz Theorem,
    Frobenius' Theorem,
    Formula for Permutations (without repetition),
    Formula for Combinations (with repetition).

\paragraph{80 Physics Theorem:} Center-of-Mass Energy,
    Planck's energy-frequency relation,
    Magnification theorem,
    Maximum Entropy Principle,
    Heron's Formula,
    Gibbs Free Energy,
    Ideal Gas Law,
    Torricelli's Law,
    Coulomb's Law,
    Gauss's Law for Electricity,
    Kirchhoff's Laws,
    Ohm's Law,
    Millma's Theorem,
    Carnot's Theorem,
    Beer-Lambert Law,
    Newton's Laws of Motion,
    Lorentz Force,
    First Law of Thermodynamics,
    Work-Energy Theorem,
    Maxwell's Equations,
    Conservation of Mechanical Energy,
    Kinetic Energy Theorem for Rotational Motion,
    Conservation of Angular Momentum,
    Torque-Angular Momentum Theorem,
    Centripetal Force Formula (for an object in circular motion),
    Euler's Rotation Theorems,
    Parallel Axis Theorem,
    Elastic Collision,
    Boucherot's Theorem (Power Factor Theorem),
    Tellegen's Theorem,
    Law of Reflection,
    Malus's Law,
    Specific Heat Capacity Formula,
    Optical Path Length (OPL) Theorem,
    Snell's Law,
    Huygens' Principle,
    Young's Double-Slit Experiment,
    Fraunhofer Diffraction Theory,
    Fresnel Equations,
    Planck's Law of Blackbody Radiation,
    Stefan-Boltzmann Law,
    Wien's Displacement Law,
    Rayleigh-Jeans Law,
    Compton Scattering Formula,
    Electric Field Formula (from a point charge),
    Speed of Sound in air (at temperature T),
    Heat Transfer via Conduction (Fourier's Law),
    Pauli Exclusion Principle,
    Energy Stored in a Capacitor Formula,
    Einstein's Photoelectric Equation,
    Bragg's Law,
    Gauss's Law for Magnetism,
    Faraday's Law of Induction,
    Lenz's Law,
    Work Done in an Adiabatic Process (Thermodynamics) Formula,
    Ampère's Circuital Law,
    Hooke's Law (for Springs),
    Laplace's Equation,
    Poisson's Equation,
    D'Alembert's Principle,
    Lagrange's Equations of Motion,
    Hamilton's Principle,
    Virial Theorem,
    Kepler's Laws of Planetary Motion,
    Newton's Law of Universal Gravitation,
    Magnetic Force on a Moving Charge Formula,
    Schwarzschild Metric,
    Lorentz Transformation,
    Einstein's Energy-Mass Equivalence,
    Shannon Entropy,
    Dirac Equation,
    Feynman Path Integral Formulation,
    Landauer's Principle,
    Onsager Reciprocity Relations,
    Bernoulli's Equation (for fluid flow),
    Stokes' Law,
    Reynolds Transport Theorem,
    Conservation of Mass,
    Thermal Conductivity Formula,
    Lens Equation (for a thin lens).

\paragraph{30 Finance Formula:}
    Binomial Model in finance,
    Net Present Value (NPV),
    Future Value (FV) Formula,
    Present Value (PV) Formula,
    Discounted Cash Flow (DCF) Model,
    Dividend Discount Model,
    Capital Asset Pricing Model (CAPM),
    Gordon Growth Model (GGM),
    Binomial Option Pricing Model,
    Bond Pricing Formula,
    Yield to Maturity (YTM),
    Sharpe Ratio,
    Macauley Duration,
    Modified Duration,
    Internal Rate of Return (IRR),
    Return on Equity (ROE),
    Value at Risk (VaR) Formula,
    Z-Spread Formula,
    Inventory Turnover Ratio,
    GDP (Expenditure Approach),
    DuPont Analysis Formula,
    Weighted Average Cost of Capital (WACC),
    Derivatives Forward Price,
    Dividend Discount Model (DDM),
    Earnings Yield Formula,
    Sustainable Growth Rate (SGR),
    Operating Leverage Formula,
    Covariance Formula,
    Variance of a Two-Asset Portfolio,
    Profitability Index (PI).

\section{Data Leakage Analysis}
\label{leak}
To ensure there is no data leakage, we compare our generated questions against the \bright queries using fuzzy string matching. We compute the longest contiguous matching subsequence and derive a similarity ratio. The highest similarity score observed is 0.5—well below the commonly used threshold of 0.8—indicating no meaningful overlap with the test data.

\section{Data Augmentation Prompts for LLM}
\label{aug_prompts}

\begin{tcolorbox}[colback=gray!10,colframe=black!75,title={\small Theorem Definition Prompt}]
\footnotesize
\setlength{\baselineskip}{1.2\baselineskip}

Your task is to provide a definition for the 
\{domain\}: \{theorem\}. Write the equation in LaTex format.\\

Here are some examples:\\
\textbf{Concept}\\
Pigeonhole Principle\\
\textbf{Definition}\\
Let $S$ be a finite set whose cardinality is $n$. Let $S_1, S_2, \ldots, S_k$ be a partition of $S$ into $k$ subsets. Then, at least one subset $S_i$ of $S$ contains at least $\left\lceil \dfrac{n}{k} \right\rceil$ elements, where $\left\lceil \cdot \right\rceil$ denotes the ceiling function.\\

Here is your task:\\
\textbf{Concept}\\
\{theorem\}\\
\textbf{Definition}\\
\end{tcolorbox}

\begin{tcolorbox}[colback=gray!10,colframe=black!75,title={\small Question Quality Check Prompt}]
\footnotesize
\setlength{\baselineskip}{1.2\baselineskip}
\textbf{Question}\\
\{question\}\\

Is this problem testing or requiring \{domain\} \{theorem\}? If yes, please answer "YES". If no, please response with a new problem and solution about \{theorem\} with similar context and difficulty. Do not provide any explanation.

\end{tcolorbox}

\begin{tcolorbox}[colback=gray!10,colframe=black!75,title={\small Solution Quality Check Prompt}]
\footnotesize
\setlength{\baselineskip}{1.2\baselineskip}
\textbf{Question}\\
\{question\}\\
\textbf{Solution}\\
\{solution\}
\\
\\Is this a correct solution to the problem and using the \{domain\} \{theorem\}? Response "YES" or "No".'

\end{tcolorbox}

\begin{tcolorbox}[colback=gray!10,colframe=black!75,title={\small Math Question-Solution Generation Prompt}]
\footnotesize
\setlength{\baselineskip}{1.2\baselineskip}

Your task is to create one \{question type\} problem with a correct solution.\\

- The problem should be new and unique, not similar to common existing problems.\\
\{``- The problem should be based on real world human activities but not a proof problem.'', ``- The problem should a multi-choice problem but not a proof problem.'', ``- The problem should be theoretical and mathematical but not a proof problem.''\}\\
- The problem should involve numerical operations.\\
- Most importantly, the problem should require or test about the \{domain\}: \{theorem\}.\\
- The problem should not explicitly mentioning {theorem}.\\
- The problem should be as difficult as \{``American Mathematics Competitions'', ``International Mathematical Olympiad'', ``Scholastic Assessment Test Math Exam''\}.\\
- The problem should be solved \{``in multiple steps'', ``by multiple {domain}s''\}.\\
\{``- The problem should be around four sentences long.'', None\}\\
\{``- The solution should not explicitly mention \{theorem\}.'', None\}\\
- The solution should include reasoning or calculation steps.\\

Write the problem after the \textbf{Problem} tag and the solution after the \textbf{Solution} tag. Do not write any explanation.
\end{tcolorbox}

\newpage
\begin{tcolorbox}[colback=gray!10,colframe=black!75,title={\small Coding Question-Solution Generation Prompt}]
\footnotesize
\setlength{\baselineskip}{1.2\baselineskip}

Your task is to create one \{question type\} problem with a correct solution.\\

- The problem should be new and unique, not similar to common existing problems.\\
\{
``- The problem should be based on real world human activities.'',
``- The problem should be based on a theoretical coding context.'',
``- The problem should be about a company or a factory.'',
``- The problem should be about a game or a puzzle.'',
``- The problem should be about designing a system.'',
``- The problem should be about a mathematical task needing automation.'',
``- The problem should be about traffic or logistics.'',
``- The problem should be about a city or a community.'',
``- The problem should be about fiance or business.'',
``- The problem should be about software or mobile applications.'',
``- The problem should be about education or e-learning.'',
``- The problem should be about e-commerce or online marketplaces.'',
``- The problem should be about agriculture or food production.'',
``- The problem should be about health or fitness.'',
``- The problem should be about customer service.'',
``- The problem should be about environmental sustainability.'',
``- The problem should be based on real world human activities.'', None\}\\
- Most importantly, the problem should require or test about the \{domain\}: \{theorem\}.\\
- The problem should be as difficult as \{``LeetCode'', ``Codeforces Contests'', ``Google Code Jam''\}.\\
- The solution code should be written in the programming language \{language\}.\\

Write the problem after the \textbf{Problem} tag and the solution after the \textbf{Solution} tag. Do not write any explanation.
\end{tcolorbox}

\newpage

\begin{tcolorbox}[colback=gray!10,colframe=black!75,title={\small Physics Question-Solution Generation Prompt}]
\footnotesize
\setlength{\baselineskip}{1.2\baselineskip}

Your task is to create one \{question type\} problem with a correct solution.\\

- The problem should be new and unique, not similar to common existing problems.\\
\{``- The problem should be based on real world human activities.'', None\}\\
- Most importantly, the problem should require or test about the \{domain\}: \{theorem\}.\\
- The problem should be as difficult as \{``International Physics Olympiad (IPhO)'', ``American Invitational Physics Exam (AIPMT)'', ``Scholastic Assessment Test (SAT) Physics Subject Test''\}.\\
- The problem should be solved \{``in multiple steps'', ``by multiple {domain}s''\}.\\
- The solution should include reasoning or calculation steps.\\

Write the problem after the \textbf{Problem} tag and the solution after the \textbf{Solution} tag. Do not write any explanation.
\end{tcolorbox}

\begin{tcolorbox}[colback=gray!10,colframe=black!75,title={\small Hard Negative Generation Prompt for LLM}]
\footnotesize
\setlength{\baselineskip}{1.2\baselineskip}

You have been assigned a retrieval task: \{instruction\} \\ \\
You will be given a user query and a positive document. Your mission is to write one hard negative document. The hard negative document must:\\
- Have the similar context background as the user query but test or require a different \{domain\}.\\
- Follow the format of the positive document.\\
- Should not be related to \{theorem\}.\\
- Should not be helpful for solving the user query problem.\\ 

\textbf{User Query}\\
\{query\}\\

\textbf{Positive Document}\\
\{pos\}\\

Directly response with the content of hard negative document.\\
\textbf{Hard Negative Document}

\end{tcolorbox}

\begin{table*}[b!]
\centering
\renewcommand{\arraystretch}{1.1}
\captionsetup{justification=justified,singlelinecheck=false}
\begin{tabularx}{\textwidth}{
    >{\raggedright\arraybackslash}l 
    >{\centering\arraybackslash}X 
    >{\centering\arraybackslash}X 
    >{\centering\arraybackslash}X
    >{\centering\arraybackslash}X
    >{\centering\arraybackslash}X
    >{\centering\arraybackslash}X
    >{\centering\arraybackslash}X
} 
\toprule
 &\textbf{\textit{N}} & \textbf{TheoT.} & \textbf{TheoQ.} & \textbf{AoPS}  & \textbf{Leet.} & {\textbf{Avg.}}    \\
\hline

Llama3     &    \multirow{7}{*}{10k}    &   31.0      &  34.6   & 5.8 &  18.9  & 22.6\\
\makecell[l]{\quad  + LLM2Vec}     &  & 28.3   &    33.8 &  12.6  &  23.6 & 24.6  \\
\makecell[l]{\quad  + SimCSE}     &  &   29.0 &   36.0   &  10.4  &  24.1  & 24.9  \\
Mistral &  & 32.4 & 33.7 & 9.4 & 20.7 & 24.1 \\
\makecell[l]{\quad  + LLM2Vec}  &   & 30.8 & 30.4 &  9.1 & 22.9 & 23.3  \\
\makecell[l]{\quad  + SimCSE}   &  & 25.5 & 32.0 & 10.3 & 23.2 & 22.8 \\

 \rowcolor{gray!20}

\ours    &    & \textbf{38.9}  & \textbf{48.3} & \textbf{15.4} & \textbf{30.0}  & \textbf{33.2} \\
\bottomrule
\end{tabularx}
\caption{Results on theorem-related reasoning-intensive retrieval (\bright) for LLM-based embedding models trained with \textbf{\theoremaug} data. We report nDCG@10 for \textit{TheoremQA} with with theorem retrieval (\textit{TheoT.}) and question retrieval (\textit{TheoQ.}), \textit{AoPS}, and \textit{LeetCode} (\textit{Leet.}). SimCSE refers to the unsupervised contrastive learning with Wikipedia sentences \citep{simcse}. \textit{N} is the amount of \theoremaug samples used for training. The best results with 10k training data are in \textbf{bold}.}
\label{simcse_compare}
\end{table*}

\newpage

\section{Implementation Details}
\label{implementation}

To ensure fair comparisons, we adopt the same model configuration for all embedding models, including the contrastive learning objective, mean pooling, bidirectional attention masking, cosine similarity, and other settings. 

For training with Public E5 and \theoremaug datasets, we follow the configuration of LLM2Vec \citep{llm2vec}, while for the training with MS MARCO with Instructions dataset, we adopt the setup in Promptriever \citep{promptriever}.

\subsection{Training Details}
For all experiments, we use language models with a bidirectional attention mask, mean pooling, and contrastive learning objectives. Training is conducted on 4 A6000 GPUs using LoRA with the specified ranks. For models with MNTP, we initialize the models from pre-trained MNTP checkpoints provided by \citet{llm2vec}.

\paragraph{MS MARCO with Instructions.} We train for 1 epoch using a batch size of 8 per GPU, 4 gradient accumulation steps, a learning rate of 1e-4 with 20 warmup steps, and LoRA rank 32. Only explicit negatives are used. Maximum input and document lengths are 304 and 196 tokens, respectively.

\paragraph{Public E5 and \theoremaug.} We train for 1 epoch with a batch size of 4 per GPU, 1 gradient accumulation step, a learning rate of 1e-4 with 100 warmup steps, and LoRA rank 16. Both hard negatives and in-batch negatives are employed. Maximum input and document lengths are set to 4096 tokens.

\subsection{Evaluation Metrics}

We use the MTEB \citep{mteb} implementations to evaluate \longembed, \bright, \followir, and general text embedding benchmarks (e.g., \Cref{general_text_embed}). The reported scores follow the main evaluation metrics defined in the original benchmark papers: Acc@1 for synthetic tasks in \longembed, nDCG@10 for real tasks in \longembed and all tasks in \bright, nDCG@5 for News21 in \followir, MAP@1000 for Robust04 and Core17 in \followir, and $p$-MRR for all subsets in \followir. The definitions of these metrics refer to original papers.

Task instructions used in \bright are as follows:
\begin{itemize}
    \item \textbf{TheoT.}: Given a problem, retrieve the relevant theorems that help solve the given problem.
    \item \textbf{TheoQ.}: Given a problem, retrieve the relevant problems that can be solved by the similar theorems.
    \item \textbf{AoPS}: Given a problem, rretrieve the relevant problems that can be solved by the similar math theorems.
    \item \textbf{Leet.}: Given a coding problem, retrieve the relevant problems that can be solved by the similar algorithms or data structures.
\end{itemize}
We follow LLM2Vec \citep{llm2vec} for other task instructions. 

\newpage
\section{Unsupervised Contrastive Learning}
\label{sec: simcse}
We conduct additional experiments for three base models to evaluate whether unsupervised learning methods, such as SimCSE \citep{simcse}, enhance retrieval performance on reasoning-intensive tasks like \bright \citep{bright}. As shown in \Cref{simcse_compare}, no significant improvement is observed, consistent with findings by \citet{llm2vec}. This can be attributed to the fundamental mismatch between semantic similarity objectives in unsupervised sentence embeddings and the theorem-centric nature of \bright, which demands deeper reasoning beyond surface-level semantics. Therefore, we exclude these results in the main paper.

\begin{table}[b!]
\centering
\renewcommand{\arraystretch}{1.1}
\captionsetup{justification=justified,singlelinecheck=false}
\begin{tabularx}{0.48\textwidth}{
    >{\raggedright\arraybackslash}X
    >{\centering\arraybackslash}X
    >{\centering\arraybackslash}X 
} 
\toprule
 & \textbf{Leet.} & \textbf{TheoQ.} \\
\midrule
\rowcolor{gray!15}
\multicolumn{3}{l}{\textbf{Overall Statistics}} \\
\makecell[l]{Total number\\of queries} & 142 & 194 \\
\midrule
\rowcolor{gray!15}
\multicolumn{3}{l}{\textbf{nDCG@10}} \\
\ours & 30.0 & 48.3 \\
E5-Mistral & 28.7 & 26.1 \\
\midrule
\rowcolor{gray!15}
\multicolumn{3}{l}{\textbf{Gold Passage Retrieval}} \\
\makecell[l]{\ours found \\ at least one gold} & 74 & 124 \\
\makecell[l]{E5-Mistral found  \\ at least one gold} & 73 & 64 \\
\makecell[l]{Both found at least\\one same gold} & 53 & 56 \\
\midrule
\rowcolor{gray!15}
\multicolumn{3}{l}{\textbf{Disjoint Gold Passage   Found}} \\
\makecell[l]{\ours found \\ E5-Mistral failed} & 19 & 67 \\
\makecell[l]{E5-Mistral found \\ \ours failed} & 18 & 7 \\
\bottomrule
\end{tabularx}
\caption{Comparative analysis of top 10 documents retrieved by \ours and E5-Mistral models on Leet. and TheoQ. subsets of \bright.}
\label{count_comparison}
\end{table}

\section{Case Study}
\label{case_study}

To further analyze models' behaviors, we collect statistics on the top 10 retrieved documents from \ours and E5-Mistral for the Leet. and TheoQ. tasks. As shown in \Cref{count_comparison}, while both models perform similarly on Leet., each model retrieves the correct document in approximately 70 cases, but there are about 20 cases where the other model fails, suggesting the presence of multiple relevance definitions in Leet. This discrepancy is less pronounced in TheoQ., indicating more consistent relevance annotations. Notably, \ours demonstrates more significant improvements on TheoQ. and TheoT..

Qualitative analysis, detailed in \Cref{case_1,case_2,case_3}, provides further insights: \Cref{case_1} reveals that E5-Mistral prioritizes semantic matching over theorem matching; \Cref{case_2} presents a failure case for E5-Mistral in identifying the correct algorithm; \Cref{case_3} illustrates a failure case for \ours, where a noisy document receives a high score due to the relevance of its first half. Identifying inconsistencies within a single document presents a new challenge for embedding models.

\begin{table}[b!]
\centering
\renewcommand{\arraystretch}{1.1}
\captionsetup{justification=justified,singlelinecheck=false}
\begin{tabularx}{0.4\textwidth}{
    >{\raggedright\arraybackslash}X
    >{\centering\arraybackslash}p{0.6cm}
    >{\centering\arraybackslash}p{0.6cm}
    >{\centering\arraybackslash}X 
} 
\toprule
 & $\checkmark$ & $\times$ & Error Rate\\
\midrule
\textbf{TheoQ.} & 14 & 4 & 22\% \\
\textbf{AoPS} & 9 & 12 & 57\%\\
\textbf{Leet.} & 10 & 8 & 44\%\\
\bottomrule
\end{tabularx}
\caption{Human re-evaluation of gold evidence in \bright\ subsets. For each subset (TheoQ., AoPS, Leet.), we manually inspect the gold documents associated with the first 10 test examples. A gold document is marked with a checkmark ($\checkmark$) if it uses a theorem or algorithm similar to that in the query, and with a cross ($\times$) otherwise. We report the proportion of incorrect gold documents as the error rate. Higher error rates are observed in AoPS and Leet. compared to TheoQ..}
\label{human_reeval}
\end{table}

\section{Noisy Data in \bright}
\label{noise_data}

\subsection{TheoQ.}
We re-examined the gold annotations for the first ten test queries in the \bright tasks. As shown in \Cref{human_reeval}, TheoQ. exhibits a lower annotation error rate compared to Leet and AoPS. Given that TheoQ. and TheoT. originate from the same underlying dataset, we expect higher sample quality in both. Accordingly, our analysis places greater emphasis on TheoQ. and TheoT., where we conduct more detailed studies.

\subsection{AoPS}
Through manual evaluation of \ours retrieved documents in the AoPS task, we observe a significant number of relevant documents that are not annotated as gold. In some queries, this number exceeds 10, undermining the effectiveness of nDCG@10 as a primary evaluation metric. This misalignment partially explains why most embedding-based retrieval models report low nDCG@10 scores (around 10\%)—the actual performance is underestimated. Consequently, comparing models using this metric becomes unreliable. For such scenario with a large number of relevant documents in the corpus, we recommend to use the LLM-as-judge approach for evaluation like \citet{mrag,ifir}.

\subsection{Leet.}

The Leet. subset often exhibits inconsistencies between query relevance and gold annotation. A notable example is the pair Trapping Rain Water I and Trapping Rain Water II, which are regarded as relevant to each other. However, they differ significantly:
\begin{itemize}
    \item \textbf{Trapping Rain Water I} is a 1D problem solved with two pointers or a stack by tracking left and right boundaries.
    \item  \textbf{Trapping Rain Water II} is a 2D problem requiring a min-heap and BFS to simulate water flow from the lowest boundaries inward.
\end{itemize}

Such cases illustrate a deviation from the intended task goal—retrieving problems and solutions that reflect the same underlying algorithmic design. 

Additionally, we identify documents like Listing 4, where the problem statement and solution in the same document correspond to different tasks. In the situation like \Cref{case_3}, documents may include a relevant question but an irrelevant solution, yet are not marked as relevant. According to the benchmark’s definition, corpus documents should offer accurate reference demonstrations or tutorials, which is not always upheld.

\begin{table*}
\centering
\renewcommand{\arraystretch}{1.2}
\captionsetup{justification=justified,singlelinecheck=false}
\begin{tabularx}{\textwidth}{
    >{\raggedright\arraybackslash}X
} 
\toprule

\textbf{Query Question}: Mary is planning to bake exactly 10 cookies, and each cookie may be one of three different shapes -- triangle, circle, and square. Mary wants the cookie shapes to be a diverse as possible. What is the smallest possible count for the most common shape across the ten cookies?  \\
\hdashline
\textbf{Gold Theorem}: Pigeonhole principle \\
\hdashline
\textbf{\ours Retrieved Document}: \\
In a group of 1000 people, at least how many people have to share the same birthday? \\
Using the \textbf{\textcolor{ForestGreen}{Pigeonhole Principle}}, we can determine the minimum number of people who must share the same birthday. In this case, the ``pigeons'' are the 1000 people, and the ``holes'' are the 365 possible birthdays (ignoring leap years) ...\\ 
\hdashline
\textbf{E5-Mistral Retrieved Document}: \\
Let $n$ represent the \textbf{\textcolor{Maroon}{smallest}} integer that satisfies the following conditions:\\
$\frac n2$ is a perfect \textbf{\textcolor{Maroon}{square}}.\\
$\frac n3$ is a perfect cube.\\
$\frac n5$ is a perfect fifth.\\
How many divisors does $n$ have that are not multiples of \textbf{\textcolor{Maroon}{10}}? \\
The first condition implies that the power of each prime factor of $n$ must be an even power (excluding $2$, which must be an odd power). The second condition implies that the power of each prime factor of $n$ must be divisible by $3$ ... \\
\bottomrule
\end{tabularx}
\caption{Qualitative comparison on \bright TheoremQA Questions: \ours retrieves a question related to the gold theorem ``Pigeonhole Principle'', whereas the most relevant question retrieved by E5-Mistral is about ``Least Common Multiple'', highlighting that E5-Mistral tends to prioritize semantic matching than theorem matching.}
\label{case_1}
\end{table*}

\begin{figure*}[h] 
\begin{lstlisting}[language=Python, caption={\ours Retrieved Code Document from from BRIGHT}]
def fallingSquares(positions):
    """
    There are several squares being dropped onto the X-axis of a 2D plane.

    You are given a 2D integer array `positions` where `positions[i] = [lefti, sideLengthi]` represents the `ith` square with a side length of `sideLengthi` that is dropped with its left edge aligned with X-coordinate `lefti`.

    Each square is dropped one at a time from a height above any landed squares. It then falls downward (negative Y direction) until it either lands **on the top side of another square** or **on the X-axis**. A square brushing the left/right side of another square does not count as landing on it. Once it lands, it freezes in place and cannot be moved.
    
    After each square is dropped, you must record the **height of the current tallest stack of squares**.
    
    Return _an integer array_ `ans` _where_ `ans[i]` _represents the height described above after dropping the_ `ith` _square_.
    
    **Example 1:**
    
    **Input:** positions = \[\[1,2\],\[2,3\],\[6,1\]\]
    **Output:** \[2,5,5\]
    **Explanation:**
    After the first drop, the tallest stack is square 1 with a height of 2.
    After the second drop, the tallest stack is squares 1 and 2 with a height of 5.
    After the third drop, the tallest stack is still squares 1 and 2 with a height of 5.
    Thus, we return an answer of \[2, 5, 5\].
    
    **Example 2:**
    
    **Input:** positions = \[\[100,100\],\[200,100\]\]
    **Output:** \[100,100\]
    **Explanation:**
    After the first drop, the tallest stack is square 1 with a height of 100.
    After the second drop, the tallest stack is either square 1 or square 2, both with heights of 100.
    Thus, we return an answer of \[100, 100\].
    Note that square 2 only brushes the right side of square 1, which does not count as landing on it.
    
    **Constraints:**
    
    *   `1 <= positions.length <= 1000`
    *   `1 <= lefti <= 108`
    *   `1 <= sideLengthi <= 106`
    """

    ans = []
    intervals = []

    for p in positions:
        L, size = p
        R = L + size
        h = size
        for h2, R2 in intervals:
            if R2 > L and R > R2:
                h = max(h, size + h2)

        maxHeight = max((h2 for h2, R2 in intervals), default=0)
        ans.append(max(maxHeight, h))
        intervals.append((h, R))

    return ans
    
\end{lstlisting}
\end{figure*}

\begin{figure*}[h] 
\begin{lstlisting}[language=Python, caption={E5-Mistral Retrieved Code Document from from BRIGHT}]

def countSmaller(nums):
    """
    Given an integer array `nums`, return _an integer array_ `counts` _where_ `counts[i]` _is the number of smaller elements to the right of_ `nums[i]`.
    
    **Example 1:**
    
    **Input:** nums = \[5,2,6,1\]
    **Output:** \[2,1,1,0\]
    **Explanation:**
    To the right of 5 there are **2** smaller elements (2 and 1).
    To the right of 2 there is only **1** smaller element (1).
    To the right of 6 there is **1** smaller element (1).
    To the right of 1 there is **0** smaller element.
    
    **Example 2:**
    
    **Input:** nums = \[-1\]
    **Output:** \[0\]
    
    **Example 3:**
    
    **Input:** nums = \[-1,-1\]
    **Output:** \[0,0\]
    
    **Constraints:**
    
    *   `1 <= nums.length <= 105`
    *   `-104 <= nums[i] <= 104`
    """

    def merge_sort(indices):
        if len(indices) <= 1:
            return indices
        mid = len(indices) // 2
        left = merge_sort(indices[:mid])
        right = merge_sort(indices[mid:])
        return merge(left, right)

    def merge(left, right):
        merged, count = [], 0
        while left and right:
            if nums[left[0]] <= nums[right[0]]:
                counts[left[0]] += count
                merged.append(left.pop(0))
            else:
                count += len(left)
                merged.append(right.pop(0))
        for i in left:
            counts[i] += count
        return merged + left + right

    counts = [0] * len(nums)
    merge_sort(list(range(len(nums))))
    return counts
    
\end{lstlisting}
\end{figure*}

\begin{table*}
\centering
\renewcommand{\arraystretch}{1.2}
\captionsetup{justification=justified,singlelinecheck=false}
\begin{tabularx}{\textwidth}{
    >{\raggedright\arraybackslash}X
} 
\toprule

\textbf{Query Question}: \\
A city's skyline is the outer contour of the silhouette formed by all the buildings in that city when viewed from a distance. Given the locations and heights of all the buildings, return \_the skyline formed by these buildings collectively\_. \\

The geometric information of each building is given in the array \texttt{buildings} where \texttt{buildings[i] = [lefti, righti, heighti]}: \\

*   \texttt{lefti} is the x coordinate of the left edge of the \texttt{ith} building. \\
*   \texttt{righti} is the x coordinate of the right edge of the \texttt{ith} building. \\
*   \texttt{heighti} is the height of the \texttt{ith} building. \\

You may assume all buildings are perfect rectangles grounded on an absolutely flat surface at height \texttt{0}. The skyline should be represented as a list of ``key points'' sorted by their x-coordinate in the form \texttt{[[x1,y1],[x2,y2],...]}. Each key point is the left endpoint of some horizontal segment in the skyline except the last point in the list, which always has a y-coordinate \texttt{0} and is used to mark the skyline's termination where the rightmost building ends. Any ground between the leftmost and rightmost buildings should be part of the skyline's contour.
\\
Note: There must be no consecutive horizontal lines of equal height in the output skyline. For instance, \texttt{[...,[2 3],[4 5],[7 5],[11 5],[12 7],...]} is not acceptable; the three lines of height 5 should be merged into one in the final output as such: \texttt{[...,[2 3],[4 5],[12 7],...]}\\
...  \\
\hdashline
\textbf{Gold Theorem}: Sweep Line Algorithm \\
\hdashline
\textbf{\ours Retrieved Document}: Listing 1\\
\hdashline
\textbf{E5-Mistral Retrieved Document}: Listing 2 \\
\bottomrule
\end{tabularx}
\caption{Qualitative comparison on \bright Leetcode: \ours retrieves code related to the gold theorem ``Sweep Line Algorithm'' (see Listing 1), while the most relevant code retrieved by E5-Mistral pertains to``Merge Sort'' (see Listing 2).}
\label{case_2}
\end{table*}

\begin{table*}
\centering
\renewcommand{\arraystretch}{1.2}
\captionsetup{justification=justified,singlelinecheck=false}
\begin{tabularx}{\textwidth}{
    >{\raggedright\arraybackslash}X
} 
\toprule

\textbf{Query Question}: \\
Convert a non-negative integer \texttt{num} to its English words representation. \\
Example 1:\\
Input: num = 123\\
Output: One Hundred Twenty Three\\
Example 2:\\
Input: num = 12345\\
Output: Twelve Thousand Three Hundred Forty Five\\
Example 3:\\
Input: num = 1234567\\
Output: One Million Two Hundred Thirty Four Thousand Five Hundred Sixty Seven\\
...  \\\hdashline
\textbf{Gold Theorem}: Recursive Decomposition and Mapping \\\hdashline
\textbf{E5-Mistral Retrieved Document}: Listing 3\\\hdashline
\textbf{\ours Retrieved Document}: Listing 4\\

\bottomrule
\end{tabularx}
\caption{Qualitative comparison on \bright Leetcode: E5-Mistral retrieves the gold document (see Listing 3), which is closely related to ``Mapping'', though not directly about ``Recursive Decomposition''. In contrast, \ours retrieves a non-gold document (see Listing 4) that includes a question requiring both ``Recursive Decomposition'' and ``Mapping'' but with an irrelevant solution code. All the embedding models in this work were not trained to assess the correctness of the solutions, which points out be a new dimension for data augmentation. We suspect these types of documents represent noisy data, as the corpus in \bright's setup is intended to consist of reference documents, which are expected to be correct.}
\label{case_3}
\end{table*}

\begin{figure*}[h] 
\begin{lstlisting}[language=Python, caption={E5-Mistral Retrieved Code Document from BRIGHT}]
def int_to_roman(num):
    """
    Roman numerals are represented by seven different symbols: `I`, `V`, `X`, `L`, `C`, `D` and `M`.
    
    **Symbol**       **Value**
    I             1
    V             5
    X             10
    L             50
    C             100
    D             500
    M             1000
    
    For example, `2` is written as `II` in Roman numeral, just two one's added together. `12` is written as `XII`, which is simply `X + II`. The number `27` is written as `XXVII`, which is `XX + V + II`.
    
    Roman numerals are usually written largest to smallest from left to right. However, the numeral for four is not `IIII`. Instead, the number four is written as `IV`. Because the one is before the five we subtract it making four. The same principle applies to the number nine, which is written as `IX`. There are six instances where subtraction is used:
    
    *   `I` can be placed before `V` (5) and `X` (10) to make 4 and 9.
    *   `X` can be placed before `L` (50) and `C` (100) to make 40 and 90.
    *   `C` can be placed before `D` (500) and `M` (1000) to make 400 and 900.
    
    Given an integer, convert it to a roman numeral.
    
    **Example 1:**
    
    **Input:** num = 3
    **Output:**  "III "
    **Explanation:** 3 is represented as 3 ones.
    
    **Example 2:**
    
    **Input:** num = 58
    **Output:**  "LVIII "
    **Explanation:** L = 50, V = 5, III = 3.
    
    **Example 3:**
    
    **Input:** num = 1994
    **Output:**  "MCMXCIV "
    **Explanation:** M = 1000, CM = 900, XC = 90 and IV = 4.
    
    **Constraints:**
    
    *   `1 <= num <= 3999`
    """
    
        romans = [
            (1000, "M"), (900, "CM"), (500, "D"),
            (400, "CD"), (100, "C"), (90, "XC"),
            (50, "L"), (40, "XL"), (10, "X"), 
            (9, "IX"), (5, "V"), (4, "IV"), (1, "I")
        ]
        roman = ""
        for value, symbol in romans:
            while num >= value:
                roman += symbol
                num -= value
        return roman
        
    
\end{lstlisting}
\end{figure*}

\begin{figure*}[h] 
\begin{lstlisting}[language=Python, caption={\ours Retrieved Code Document from BRIGHT}]
import heapq
"""
Given an array of `digits` which is sorted in **non-decreasing** order. You can write numbers using each `digits[i]` as many times as we want. For example, if `digits = ['1','3','5']`, we may write numbers such as `'13'`, `'551'`, and `'1351315'`.

Return _the number of positive integers that can be generated_ that are less than or equal to a given integer `n`.

**Example 1:**

**Input:** digits = \[ "1 ", "3 ", "5 ", "7 "\], n = 100
**Output:** 20
**Explanation:** 
The 20 numbers that can be written are:
1, 3, 5, 7, 11, 13, 15, 17, 31, 33, 35, 37, 51, 53, 55, 57, 71, 73, 75, 77.

**Example 2:**

**Input:** digits = \[ "1 ", "4 ", "9 "\], n = 1000000000
**Output:** 29523
**Explanation:** 
We can write 3 one digit numbers, 9 two digit numbers, 27 three digit numbers,
81 four digit numbers, 243 five digit numbers, 729 six digit numbers,
2187 seven digit numbers, 6561 eight digit numbers, and 19683 nine digit numbers.
In total, this is 29523 integers that can be written using the digits array.

**Example 3:**

**Input:** digits = \[ "7 "\], n = 8
**Output:** 1

**Constraints:**

*   `1 <= digits.length <= 9`
*   `digits[i].length == 1`
*   `digits[i]` is a digit from `'1'` to `'9'`.
*   All the values in `digits` are **unique**.
*   `digits` is sorted in **non-decreasing** order.
*   `1 <= n <= 109`
"""


def minRefuelStops(target: int, startFuel: int, stations: List[List[int]]) -> int:
    i, stops, curFuel = 0, 0, startFuel
    pq = []
    while curFuel < target:
        while i < len(stations) and stations[i][0] <= curFuel:
            heapq.heappush(pq, -stations[i][1])
            i += 1
        if not pq: return -1
        curFuel += -heapq.heappop(pq)
        stops += 1
    return stops

\end{lstlisting}
\end{figure*}

\section{\theoremaug Examples}
\label{theorem_ex}

We illustrate four retrieval examples:
\begin{itemize}
    \item Math: Question-to-Theorem (\Cref{example_math_p2i})
    \item Math: Question-to-Question (\Cref{example_math_p2p})
    \item Physics: Question-to-Theorem (\Cref{example_physics_p2i})
    \item Physics: Question-to-Question (\Cref{example_physics_p2p})
\end{itemize}

\begin{table*}
\centering
\renewcommand{\arraystretch}{1.2}
\captionsetup{justification=justified,singlelinecheck=false}
\begin{tabularx}{\textwidth}{>{\raggedright\arraybackslash}X}
\toprule

\textbf{Retrieval Task}: \\
Given a math problem, retrieve the relevant math theorem that helps solve the given problem. \\\hdashline

\textbf{User Query}: \\
In a garden shaped like a right triangle, one leg measures 24 meters and the other leg measures 10 meters. If a gardener wants to install a diagonal pathway that spans the triangle's hypotenuse, how many meters of paving stones will the gardener need? A) 20 m \quad B) 30 m \quad C) 26 m \quad D) 34 m \\\hdashline

\textbf{Positive Document}: \\
Pythagorean Theorem \\
In a right-angled triangle, the square of the length of the hypotenuse (the side opposite the right angle) is equal to the sum of the squares of the lengths of the other two sides. This relationship can be expressed mathematically as: \\
\[
c^2 = a^2 + b^2
\]
where \( c \) is the length of the hypotenuse, and \( a \) and \( b \) are the lengths of the other two sides. \\\hdashline

\textbf{Hard Negative Document}: \\
Triangle Inequality \\
For any triangle with sides of lengths \( a \), \( b \), and \( c \), the following inequalities hold: \\
\[
a + b > c, \quad a + c > b, \quad \text{and} \quad b + c > a.
\]
This means that the sum of the lengths of any two sides of a triangle must be greater than the length of the remaining side. \\

\bottomrule
\end{tabularx}
\caption{\theoremaug Math Question-to-Concept Example.}
\label{example_math_p2i}
\end{table*}

\begin{table*}
\centering
\renewcommand{\arraystretch}{1.2}
\captionsetup{justification=justified,singlelinecheck=false}
\begin{tabularx}{\textwidth}{>{\raggedright\arraybackslash}X}
\toprule

\textbf{Retrieval Task}: \\
Given a math problem, retrieve the relevant problems that can be solved by the similar math theorem. \\
\hdashline

\textbf{User Query}: \\
A gardener is laying out a triangular flower bed in the backyard. The two sides of the triangle are 24 feet and 32 feet long. To ensure that the flower bed is perfectly shaped, the gardener wants to find the length of the fencing needed for the third side, which will form a right triangle at one of the corners. What is the length of the third side of the flower bed?
 \\
\hdashline

\textbf{Positive Document}: \\

A rescue team is stationed at a point 450 meters from the base of a cliff. To reach the top of the cliff, they need to scale it by climbing a rope that is secured at the top and stretches down to the base. The climber ascends along the rope, which makes an angle with the ground. If the height of the cliff is 600 meters, what is the length of the rope in meters that the climber must use?
\\
We can visualize this situation as a right triangle, where one leg represents the height of the cliff, the other leg represents the horizontal distance from the base of the cliff to the team’s position, and the hypotenuse represents the length of the rope.
\\
Let \( r \) be the length of the rope. According to the relationship:
\[
r^2 = (450)^2 + (600)^2
\]
...\\
Taking the square root:
\[
r = \sqrt{562500} = 750
\]

Thus, the length of the rope the climber must use is \( \boxed{750} \) meters.\\
 \\\hdashline

\textbf{Hard Negative Document}: \\

A gardener is designing a rectangular flower bed that has an area of 120 square meters. He wants the length of the bed to be 5 meters more than twice the width. What will be the length of the flower bed in meters? \quad A) 15 \quad B) 20 \quad C) 25 \quad D) 30\\

Let the width of the flower bed be \( w \) meters.  Then, the length \( l \) will be \( 2w + 5 \) meters. The area of the rectangle can be expressed as:
\[
l \times w = 120
\]
Substituting for \( l \):
\[
(2w + 5)w = 120
\]
This expands to:
\[
2w^2 + 5w - 120 = 0
\]
Now, we can identify the coefficients: \( a = 2 \), \( b = 5 \), and \( c = -120 \). 
Using the quadratic formula \( w = \frac{-b \pm \sqrt{b^2 - 4ac}}{2a} \):
\[
b^2 - 4ac = 5^2 - 4 \cdot 2 \cdot (-120) = 25 + 960 = 985
\]
Thus,
\[
w = \frac{-5 \pm \sqrt{985}}{4}
\]
...
\\
Thus, the correct length is \( \boxed{20} \).\\

\bottomrule
\end{tabularx}
\caption{\theoremaug Math Question-to-Question Example.}
\label{example_math_p2p}
\end{table*}

\begin{table*}
\centering
\renewcommand{\arraystretch}{1.2}
\captionsetup{justification=justified,singlelinecheck=false}
\begin{tabularx}{\textwidth}{>{\raggedright\arraybackslash}X}
\toprule

\textbf{Retrieval Task}: \\
Given a physics problem, retrieve the relevant physics theorem that help solve the given problem. \\\hdashline

\textbf{User Query}: \\
A laboratory experiment involves a cylindrical conductor made of an unknown material that has a length of 2.5 meters and a diameter of 0.01 meters. This conductor is connected to a circuit powered by a stable voltage source of 12 volts. During an experiment, the current flowing through the conductor is measured to be 0.5 amperes. Calculate the resistance of the conductor in ohms, given that the resistivity of the material is not known but can be derived from the current and voltage provided. \\\hdashline

\textbf{Positive Document}: \\
Ohm's Law\\
Ohm's Law states that the current flowing through a conductor between two points is directly proportional to the voltage across the two points and inversely proportional to the resistance of the conductor. This relationship is fundamental in electrical circuits and can be expressed mathematically as:

\[
V = I \cdot R
\]

where \( V \) is the voltage (in volts), \( I \) is the current (in amperes), and \( R \) is the resistance (in ohms).
 \\\hdashline

\textbf{Hard Negative Document}: \\
Energy Stored in a Capacitor Formula\\
The energy \( U \) stored in a capacitor is given by the formula:
\[
U = \frac{1}{2} C V^2
\]
where \( U \) is the energy (in joules), \( C \) is the capacitance (in farads), and \( V \) is the voltage across the capacitor (in volts). This equation describes how the energy stored in the electric field of a capacitor is related to its capacitance and the voltage applied across it.
 \\

\bottomrule
\end{tabularx}
\caption{\theoremaug Physics Question-to-Concept Example.}
\label{example_physics_p2i}
\end{table*}

\begin{table*}
\centering
\renewcommand{\arraystretch}{1.2}
\captionsetup{justification=justified,singlelinecheck=false}
\begin{tabularx}{\textwidth}{>{\raggedright\arraybackslash}X}
\toprule

\textbf{Retrieval Task}: \\
Given a physics problem, retrieve the problems that can be solved by the similar physics theorem. \\\hdashline

\textbf{User Query}: \\
A circuit consists of a 12V battery connected in series with two resistors: R1 and R2. R1 has a resistance of 4 ohms, and R2 has a resistance of 6 ohms. Additionally, there is a variable resistor (R3) connected in parallel with R2, which can vary its resistance. When R3 is set to 3 ohms, the total current in the circuit is measured to be 1.2A. If R3 is adjusted to 8 ohms, what is the new total current flowing through the circuit? \\\hdashline

\textbf{Positive Document}: \\

A 10-meter-long copper wire has a uniform cross-sectional area of 1.5 mm² and is used to connect a 9V battery to a small device. The resistivity of copper is \( 1.68 \times 10^{-8} \, \Omega \cdot m \). What is the current flowing through the device if the total resistance of the wire is considered, and the device has an internal resistance of 2 ohms?\\

First, we calculate the resistance of the copper wire using the formula:
\[
R_{\text{wire}} = \frac{\rho L}{A}
\]
where \( \rho \) is resistivity, \( L \) is length of the wire, and \( A \) is cross-sectional area.
\\
...
\\
Now use Ohm's law to find the current flowing through the device:
\[
I = \frac{V}{R_{\text{total}}}
\]
Substituting the values:
\[
I = \frac{9V}{2.112 \, \Omega} \approx 4.26 \, A
\]

Thus, the current flowing through the device is approximately \( 4.26 \, A \).
 \\\hdashline

\textbf{Hard Negative Document}: \\
An engineer is testing a prototype of a magnetic levitation train system. As the train approaches a section with a solenoid that is turned off, its speed is 15 m/s. The solenoid has an area of 0.8 m² and the magnetic field it produces, when activated, is 0.4 T. The train is designed so that when it approaches, an induced current is created in the conductive loops on the train due to a change in magnetic flux. The solenoid is activated precisely as the train is 2 m away from its entrance. Calculate the induced current (I) in the loop of the train if the resistance of the loop is 5 ohms. Treat the magnetic field as uniform.
\\
First, calculate the time \( t \) it takes for the train to reach the solenoid:

\[
t = \frac{d}{v} = \frac{2 \, \text{m}}{15 \, \text{m/s}} = \frac{2}{15} \, \text{s} \approx 0.1333 \, \text{s}
\]

Now, calculate the change in magnetic flux \( \Delta \Phi \) through the loops on the train as the solenoid is activated:
\\
...\\

\bottomrule
\end{tabularx}
\caption{\theoremaug Physics Question-to-Question Example.}
\label{example_physics_p2p}
\end{table*}

\end{document}